\pdfoutput=1
\documentclass{article}

\PassOptionsToPackage{numbers, compress}{natbib}
\usepackage[dblblindworkshop, final]{neurips_2026}
\workshoptitle{SLMs for Agentic Systems, Paris, France, 2026}

\usepackage[utf8]{inputenc}
\usepackage[T1]{fontenc}
\usepackage{url}
\usepackage{booktabs}
\usepackage{amsmath}
\usepackage{amssymb}
\usepackage{amsfonts}
\usepackage{graphicx}
\usepackage{nicefrac}
\usepackage{microtype}
\usepackage{tikz}       %
\usetikzlibrary{arrows.meta, positioning, fit}
\usepackage{xcolor}   %
\usepackage{hyperref}
\hypersetup{pdftitle={Do Small Language Models Learn to Negotiate? A Controlled Scaling Study of RL-Trained Sellers}, pdfauthor={Pedro Tabacof, Sagar Joglekar}}

\newcommand{\Ut}{\tilde{U}}
\newcommand{\Us}{S}
\newcommand{\ci}[2]{{\scriptsize\mbox{[#1, #2]}}}
\newcommand{\cis}[2]{{\scriptsize[#1,#2]}}

\newcommand{\gainAllEtwo}{$+0.001$ \ci{-0.019}{+0.021}}
\newcommand{\gainAllEtwoPt}{$+0.001$}
\newcommand{\gainCarEtwo}{$+0.006$ \ci{-0.028}{+0.039}}
\newcommand{\contAllEtwoPt}{$-0.004$}
\newcommand{\gainAllEfour}{$+0.023$ \ci{+0.011}{+0.035}}
\newcommand{\gainAllEfourPt}{$+0.023$}
\newcommand{\gainCarEfour}{$-0.003$ \ci{-0.045}{+0.039}}
\newcommand{\contAllEfourPt}{$+0.015$}
\newcommand{\gainAllTwelve}{$+0.063$ \ci{+0.052}{+0.075}}
\newcommand{\gainCarTwelve}{$+0.032$ \ci{+0.003}{+0.063}}
\newcommand{\contAllTwelvePt}{$+0.038$}
\newcommand{\gainAllThirtyOne}{$+0.078$ \ci{+0.064}{+0.092}}
\newcommand{\gainAllThirtyOnePt}{$+0.078$}
\newcommand{\gainCarThirtyOne}{$+0.008$ \ci{-0.019}{+0.037}}
\newcommand{\contAllThirtyOnePt}{$+0.055$}
\newcommand{\earlyFlipAll}{$-0.024$ \ci{-0.038}{-0.010}}
\newcommand{\earlyFlipCar}{$+0.019$ \ci{-0.017}{+0.058}}
\newcommand{\frontierAll}{$+0.011$ \ci{-0.001}{+0.023}}
\newcommand{\frontierCar}{$+0.068$ \ci{+0.028}{+0.107}}
\newcommand{\frontierTrain}{$+0.004$ \ci{-0.010}{+0.018}}
\newcommand{\diffEfourEtwo}{$+0.022$ \ci{-0.002}{+0.046}}
\newcommand{\diffEfourEtwoPt}{$+0.022$}
\newcommand{\diffTwelveEfour}{$+0.040$ \ci{+0.024}{+0.056}}
\newcommand{\diffTwelveEfourPt}{$+0.040$}
\newcommand{\diffThirtyOneTwelve}{$+0.015$ \ci{+0.001}{+0.029}}
\newcommand{\diffThirtyOneTwelvePt}{$+0.015$}
\newcommand{\tunedAllEtwo}{$+0.037$ \ci{+0.017}{+0.057}}
\newcommand{\tunedAllEtwoPt}{$+0.037$}
\newcommand{\tunedCarEtwo}{$+0.059$ \ci{+0.023}{+0.095}}
\newcommand{\tunedVsStepEtwoPt}{$+0.032$}
\newcommand{\tunedVsContEtwo}{$+0.036$ \ci{+0.018}{+0.054}}
\newcommand{\tunedAllEfour}{$+0.089$ \ci{+0.075}{+0.102}}
\newcommand{\tunedAllEfourPt}{$+0.089$}
\newcommand{\tunedCarEfour}{$+0.042$ \ci{-0.007}{+0.091}}
\newcommand{\tunedVsStepEfourPt}{$+0.081$}
\newcommand{\tunedVsContEfour}{$+0.065$ \ci{+0.053}{+0.079}}
\newcommand{\tunedCarTwelve}{$+0.020$ \ci{-0.021}{+0.061}}
\newcommand{\tunedCarThirtyOne}{$+0.006$ \ci{-0.030}{+0.040}}
\newcommand{\tunedVsSonnetEfour}{$+0.003$ \ci{-0.013}{+0.019}}
\newcommand{\tunedVsTerraEfour}{$+0.014$ \ci{-0.003}{+0.031}}
\newcommand{\tunedVsSonnetTwelve}{$+0.086$ \ci{+0.068}{+0.104}}
\newcommand{\tunedVsTerraTwelve}{$+0.097$ \ci{+0.078}{+0.116}}
\newcommand{\tunedStepThirtyOneTwelve}{$+0.021$ \ci{+0.003}{+0.040}}
\newcommand{\tunedStepThirtyOneTwelvePt}{$+0.021$}
\newcommand{\efourStepCheck}{$-0.003$ \ci{-0.015}{+0.008}}
\newcommand{\pushedGain}{$+0.050$ \ci{+0.028}{+0.073}}
\newcommand{\pushedGainPt}{$+0.050$}
\newcommand{\pushedVsTuned}{$+0.013$ \ci{-0.010}{+0.035}}

\newcommand{\armTwelveBasePt}{$0.525$}
\newcommand{\armThirtyOneBasePt}{$0.542$}
\newcommand{\armThirtyOneRlPt}{$0.620$}
\newcommand{\armSonnetPt}{$0.515$}
\newcommand{\armTerraPt}{$0.504$}
\newcommand{\armEtwoTunedPt}{$0.460$}
\newcommand{\armEfourTunedPt}{$0.518$}
\newcommand{\armTwelveTunedPt}{$0.601$}
\newcommand{\armThirtyOneTunedPt}{$0.623$}

\newcommand{\heldoutN}{128}
\newcommand{\pairedN}{1{,}152}
\newcommand{\totalEpisodes}{24,192}
\newcommand{\numArms}{21}

\newcommand{\pushedBuyerGap}{$+0.175$}
\newcommand{\mcError}{$\pm0.0001$}
\newcommand{\jsonLeak}{$91.6\%$}
\newcommand{\sonnetVsSonnetBuyer}{$0.518$}
\newcommand{\sonnetVsTerraBuyer}{$0.513$}
\newcommand{\sonnetSelfPlayN}{576}
\newcommand{\terraVsSonnetBuyer}{$0.504$}
\newcommand{\terraVsTerraBuyer}{$0.503$}
\newcommand{\pushedBadParse}{$0.0282$}
\newcommand{\tunedBadParse}{$0.0008$}
\newcommand{\splitRangeLo}{$+0.032$}
\newcommand{\splitRangeHi}{$+0.079$}

\newcommand{\openerTunedThirtyOne}{$30.3\%$}
\newcommand{\openerIncThirtyOne}{$14.4\%$}
\newcommand{\openerRatioThirtyOne}{$2.1\times$}
\newcommand{\turnsTunedThirtyOne}{$7.3$}
\newcommand{\turnsIncThirtyOne}{$10.7$}
\newcommand{\turnsCutThirtyOne}{$32\%$}
\newcommand{\welfareTunedThirtyOne}{$1.239$}
\newcommand{\welfareIncThirtyOne}{$1.202$}
\newcommand{\splitTunedThirtyOne}{$-0.002$}
\newcommand{\carVoiceTuned}{$78.1\%$}
\newcommand{\carVoiceInc}{$14.1\%$}

\newcommand{\klThirtyOneTuned}{$0.437$}
\newcommand{\klThirtyOneInc}{$0.027$}
\newcommand{\klTwelveTuned}{$0.126$}
\newcommand{\klPushed}{$0.313$}

\newcommand{\stratOpenLo}{$0.4\%$}
\newcommand{\stratOpenHi}{$26.0\%$}
\newcommand{\stratLateBaseLo}{$3.7\%$}
\newcommand{\stratLateBaseHi}{$20.0\%$}
\newcommand{\stratOpenTwelveTuned}{$76.6\%$}
\newcommand{\stratTermsTwelveTuned}{$6.6$}
\newcommand{\stratTermShareTwelveTuned}{$74\%$}
\newcommand{\stratOpenUTwelveTuned}{$0.741$}
\newcommand{\stratConcTwelveTuned}{$0.128$}
\newcommand{\stratWelfTwelveInc}{$+0.008$ \ci{-0.008}{+0.023}}
\newcommand{\stratBuyerShareTwelveInc}{$88\%$}
\newcommand{\stratWelfTwelveTuned}{$+0.035$ \ci{+0.012}{+0.057}}
\newcommand{\stratBuyerShareTwelveTuned}{$55\%$}
\newcommand{\stratOpenThirtyOneTuned}{$99.7\%$}
\newcommand{\stratTermsThirtyOneTuned}{$6.2$}
\newcommand{\stratTermShareThirtyOneTuned}{$69\%$}
\newcommand{\stratOpenUThirtyOneTuned}{$0.757$}
\newcommand{\stratConcThirtyOneTuned}{$0.135$}
\newcommand{\stratWelfThirtyOneInc}{$+0.005$ \ci{-0.013}{+0.025}}
\newcommand{\stratBuyerShareThirtyOneInc}{$93\%$}
\newcommand{\stratWelfThirtyOneTuned}{$+0.042$ \ci{+0.023}{+0.060}}
\newcommand{\stratBuyerShareThirtyOneTuned}{$48\%$}
\newcommand{\stratDistShareLo}{$75\%$}
\newcommand{\stratDistShareHi}{$81\%$}
\newcommand{\stratIntegThirtyOneInc}{$+0.016$ \ci{+0.007}{+0.024}}
\newcommand{\stratBuyerIntegThirtyOneInc}{$-0.040$ \ci{-0.051}{-0.029}}
\newcommand{\stratGapEfourInc}{$+0.012$ \ci{+0.001}{+0.024}}
\newcommand{\stratLateLo}{$-7.5$}
\newcommand{\stratLateHi}{$+5.0$}
\newcommand{\stratLatePushed}{$+13.0$}
\newcommand{\stratOpenEfourTuned}{$20.7\%$}
\newcommand{\pushedGainSonnet}{$-0.012$ \ci{-0.040}{+0.017}}
\newcommand{\pushedGainTerra}{$+0.112$ \ci{+0.084}{+0.141}}
\newcommand{\pushedSplitVsBase}{$+0.124$ \ci{+0.089}{+0.160}}

\newcommand{\negscaleTableFactorial}{%
\begin{table}[t]
  \centering
  \caption{Scores for four sizes at two learning rates over all nine domains, with pointwise 95\% seed-cluster bootstrap intervals on paired within-scenario differences. Columns 5--6 compare each 3$\times$ LR arm (read at step 56) with the shared-LR arm of the same size, first at the early checkpoints (56 v.\ 56 at 4.5B and 31B, 56 v.\ 64 at 2.3B and 12B), then with the shared-LR arm read at 104 steps, roughly twice the steps and \emph{not} compute-matched. Bold with $*$ survives Holm correction at $\alpha = 0.05$ across the prespecified primary family of four (column 3); $\mathrm{e}$ marks exploratory, unadjusted contrasts whose interval excludes zero, and columns 4--6 are exploratory throughout. $\dagger$ passes the four prespecified guards but fails two post-hoc style diagnostics (Appendix~\ref{app:degenerate}).}
  \label{tab:factorial}
  \scriptsize
  \setlength{\tabcolsep}{3pt}
  \begin{tabular}{lccccc}
    \toprule
    & & \multicolumn{2}{c}{$-$ base} & \multicolumn{2}{c}{3$\times$ LR $-$ shared LR} \\
    \cmidrule(lr){3-4} \cmidrule(lr){5-6}
    Size & 3$\times$ LR $\Us$ & shared LR & 3$\times$ LR & early ckpt (56 v.\ 56/64) & @104 steps \\
    \midrule
    2.3B         & 0.460 & $+0.001$ \cis{-0.019}{+0.021} & $+0.037^{\mathrm{e}}$ \cis{+0.017}{+0.057} & $+0.032^{\mathrm{e}}$ \cis{+0.013}{+0.051} & $+0.036^{\mathrm{e}}$ \cis{+0.018}{+0.054} \\
    4.5B         & 0.518 & $\mathbf{+0.023}^{*}$ \cis{+0.011}{+0.035} & $+0.089^{\mathrm{e}}$ \cis{+0.075}{+0.102} & $+0.081^{\mathrm{e}}$ \cis{+0.066}{+0.095} & $+0.065^{\mathrm{e}}$ \cis{+0.053}{+0.079} \\
    12B          & 0.601 & $\mathbf{+0.063}^{*}$ \cis{+0.052}{+0.075} & $+0.076^{\mathrm{e}}$ \cis{+0.059}{+0.093} & $+0.051^{\mathrm{e}}$ \cis{+0.032}{+0.068} & $+0.013$ \cis{-0.002}{+0.028} \\
    31B          & 0.623 & $\mathbf{+0.078}^{*}$ \cis{+0.064}{+0.092} & $+0.080^{\mathrm{e}}$ \cis{+0.063}{+0.098} & $+0.058^{\mathrm{e}}$ \cis{+0.041}{+0.075} & $+0.003$ \cis{-0.014}{+0.020}$^{\dagger}$ \\
    \bottomrule
  \end{tabular}
\end{table}%
}

\newcommand{\negscaleTableMain}{%
\begin{table}[h]
  \centering
  \caption{The outcome decomposition behind Figure~\ref{fig:scaling}a: base sellers, the shared-LR sellers at 104 combined optimizer steps ($10^{-6}$) and the two frontier reference sellers, $1{,}152$ paired episodes each with 95\% seed-cluster bootstrap intervals. Deal-only $\Ut$ conditions on agreement; joint welfare is the sum of both parties' normalized utilities. An arm that closes fewer but better deals can score higher on deal-only utility and lower on $\Us$.}
  \label{tab:main}
  \small
  \setlength{\tabcolsep}{5pt}
  \begin{tabular}{lccccc}
    \toprule
    Seller & Steps & $\Us$ (all domains) & Deal rate & Deal-only $\Ut$ & Joint welfare \\
    \midrule
    Gemma 4 E2B (2.3B)     & base  & 0.422 \ci{0.400}{0.446} & 0.939 & 0.469 & 1.162 \\
    Gemma 4 E4B (4.5B)     & base  & 0.429 \ci{0.407}{0.453} & 0.997 & 0.431 & 1.144 \\
    Gemma 4 12B            & base  & 0.525 \ci{0.503}{0.548} & 1.000 & 0.525 & 1.171 \\
    Gemma 4 31B            & base  & 0.542 \ci{0.518}{0.567} & 1.000 & 0.542 & 1.197 \\
    \midrule
    Gemma 4 E2B (2.3B)     & 104   & 0.424 \ci{0.397}{0.451} & 0.909 & 0.496 & 1.177 \\
    Gemma 4 E4B (4.5B)     & 104   & 0.453 \ci{0.430}{0.476} & 0.997 & 0.455 & 1.168 \\
    \textbf{Gemma 4 12B}   & \textbf{104} & \textbf{0.588} \ci{0.570}{0.608} & 0.998 & 0.590 & 1.178 \\
    \textbf{Gemma 4 31B}   & \textbf{104} & \textbf{0.620} \ci{0.597}{0.643} & 1.000 & 0.620 & 1.202 \\
    \midrule
    \midrule
    Claude Sonnet~5        & ---   & 0.515 \ci{0.492}{0.539} & 1.000 & 0.515 & 1.175 \\
    GPT-5.6-Terra          & ---   & 0.504 \ci{0.478}{0.530} & 0.999 & 0.505 & 1.211 \\
    \bottomrule
  \end{tabular}
\end{table}%
}

\newcommand{\negscaleTableGrid}{%
\begin{table}[h]
  \centering
  \caption{Every evaluated arm, $1{,}152$ paired episodes each. $\Us$ is the paired score ($\Ut$ on a deal, $-0.3$ on a no-deal) with 95\% seed-cluster bootstrap intervals; ``held-out'' is the withheld ninth domain ($n = 128$). Steps are combined optimizer steps; turns are individual turns out of the 24 a 12-round episode allows. Bad-parse here is over \emph{all} turns, seller and buyer; Table~\ref{tab:guards} reports it over the seller's own turns only, which roughly doubles it. The 10$\times$ LR arm is in Table~\ref{tab:guards} instead.}
  \label{tab:grid}
  \small
  \setlength{\tabcolsep}{4.5pt}
  \begin{tabular}{lcccccc}
    \toprule
    Seller & Steps & $\Us$ (all domains) & $\Us$ (held-out) & Deal rate & Turns & Bad-parse (all turns) \\
    \midrule
    \multicolumn{7}{l}{\emph{Gemma 4, base (no RL)}} \\
    E2B (2.3B)         & ---   & 0.422 \ci{0.400}{0.446} & 0.475 \ci{0.423}{0.527} & 0.939 & 11.9 & 0.0004 \\
    E4B (4.5B)         & ---   & 0.429 \ci{0.407}{0.453} & 0.467 \ci{0.417}{0.517} & 0.997 & 12.2 & 0.0021 \\
    12B                & ---   & 0.525 \ci{0.503}{0.548} & 0.561 \ci{0.508}{0.613} & 1.000 & 11.8 & 0.0004 \\
    31B                & ---   & 0.542 \ci{0.518}{0.567} & 0.638 \ci{0.582}{0.693} & 1.000 & 10.3 & 0.0000 \\
    \midrule
    \multicolumn{7}{l}{\emph{Gemma 4, after GRPO at the shared LR $10^{-6}$}} \\
    E2B (2.3B)         & 64    & 0.428 \ci{0.406}{0.451} & 0.478 \ci{0.432}{0.525} & 0.920 & 12.6 & 0.0007 \\
    E2B (2.3B)         & 104   & 0.424 \ci{0.397}{0.451} & 0.481 \ci{0.425}{0.537} & 0.909 & 12.8 & 0.0006 \\
    E4B (4.5B)         & 56    & 0.437 \ci{0.414}{0.461} & 0.461 \ci{0.410}{0.514} & 0.998 & 11.3 & 0.0028 \\
    E4B (4.5B)         & 104   & 0.453 \ci{0.430}{0.476} & 0.464 \ci{0.413}{0.517} & 0.997 & 11.7 & 0.0001 \\
    12B                & 64    & 0.551 \ci{0.528}{0.574} & 0.573 \ci{0.520}{0.628} & 0.995 & 12.5 & 0.0002 \\
    \textbf{12B}       & \textbf{104} & \textbf{0.588} \ci{0.570}{0.608} & 0.593 \ci{0.541}{0.644} & 0.998 & 12.3 & 0.0003 \\
    31B                & 40    & 0.563 \ci{0.540}{0.587} & 0.643 \ci{0.590}{0.695} & 1.000 & 10.2 & 0.0001 \\
    31B                & 56    & 0.564 \ci{0.540}{0.589} & 0.613 \ci{0.559}{0.667} & 1.000 &  9.9 & 0.0010 \\
    \textbf{31B}       & \textbf{104} & \textbf{0.620} \ci{0.597}{0.643} & \textbf{0.647} \ci{0.594}{0.699} & 1.000 & 10.7 & 0.0000 \\
    \midrule
    \multicolumn{7}{l}{\emph{Gemma 4, after GRPO at the 3$\times$ LR}} \\
    E2B (2.3B)         & 56    & 0.460 \ci{0.433}{0.486} & 0.534 \ci{0.482}{0.588} & 0.948 & 11.9 & 0.0004 \\
    E4B (4.5B)         & 56    & 0.518 \ci{0.496}{0.540} & 0.509 \ci{0.456}{0.562} & 1.000 & 11.4 & 0.0002 \\
    E4B (4.5B)         & 64    & 0.515 \ci{0.493}{0.537} & 0.508 \ci{0.456}{0.562} & 1.000 & 11.6 & 0.0003 \\
    12B                & 56    & 0.601 \ci{0.580}{0.623} & 0.582 \ci{0.528}{0.633} & 0.999 & 11.2 & 0.0000 \\
    31B                & 56    & 0.623 \ci{0.602}{0.643} & 0.644 \ci{0.586}{0.703} & 1.000 &  7.3 & 0.0000 \\
    \midrule
    \multicolumn{7}{l}{\emph{Frontier reference sellers (no RL)}} \\
    Claude Sonnet~5    & ---   & 0.515 \ci{0.492}{0.539} & 0.544 \ci{0.483}{0.606} & 1.000 & 10.4 & 0.0000 \\
    GPT-5.6-Terra      & ---   & 0.504 \ci{0.478}{0.530} & 0.477 \ci{0.419}{0.534} & 0.999 &  8.8 & 0.0000 \\
    \bottomrule
  \end{tabular}
\end{table}%
}

\newcommand{\negscaleTableEfficiency}{%
\begin{table}[h]
  \centering
  \caption{Score beside the hardware it took to serve in our evaluation configuration; we do not benchmark throughput, latency or cost. Accelerators are 48\,GB devices at evaluation time (Table~\ref{tab:topology}); the 31B arms need four because the KV cache for a 16k context does not fit on fewer. ``Shared LR'' is the 104-step arm at $10^{-6}$ and ``3$\times$ LR'' the $3{\times}10^{-6}$ arm read at step 56, both from Table~\ref{tab:grid}. As sellers the frontier arms each play half their episodes against a same-model buyer, which no Gemma arm does (Appendix~\ref{app:limitations}). They are API-served, so no accelerator count applies; the reference point for them is \$0.069 per episode, the mean over the four frontier-seller cells (37.1M input and 5.2M output tokens) with both sides billed at the providers' list prices at evaluation time.}
  \label{tab:efficiency}
  \small
  \setlength{\tabcolsep}{5pt}
  \begin{tabular}{lcccc}
    \toprule
    Seller & Effective / total & Accelerators & GB served & $\Us$ \\
    \midrule
    Gemma 4 E2B, shared LR           & 2.3B / 5.1B  & 1 & 48 & 0.424 \\
    Gemma 4 E4B, shared LR           & 4.5B / 8.0B  & 1 & 48 & 0.453 \\
    Gemma 4 12B, shared LR           & 11.95B       & 1 & 48 & 0.588 \\
    Gemma 4 31B, shared LR           & 30.7B        & 4 & 192 & 0.620 \\
    \midrule
    Gemma 4 E2B, 3$\times$ LR        & 2.3B / 5.1B  & 1 & 48 & 0.460 \\
    Gemma 4 E4B, 3$\times$ LR        & 4.5B / 8.0B  & 1 & 48 & 0.518 \\
    \textbf{Gemma 4 12B, 3$\times$ LR} & 11.95B       & 1 & 48 & 0.601 \\
    Gemma 4 31B, 3$\times$ LR        & 30.7B        & 4 & 192 & 0.623 \\
    \midrule
    Gemma 4 31B, base                & 30.7B        & 4 & 192 & 0.542 \\
    Claude Sonnet~5                  & ---          & --- & --- & 0.515 \\
    GPT-5.6-Terra                    & ---          & --- & --- & 0.504 \\
    \bottomrule
  \end{tabular}
\end{table}%
}

\newcommand{\negscaleTableGuards}{%
\begin{table}[h]
  \centering
  \caption{The 3$\times$ and 10$\times$ LR arms at 2.3B beside the base and the two shared-LR arms, on the prespecified deal-rate, parse and turn guards and on the post-hoc buyer split. Bad-parse is over the seller's own turns, so it is about twice the all-turn rate in Table~\ref{tab:grid}. The buyer split is $\Us$ against \mbox{GPT-5.6-Terra} minus $\Us$ against Claude Sonnet~5; the training buyers were GPT-5.5 (20\%) and \mbox{GPT-5.4-mini} (80\%), so Terra shares their model family and Sonnet~5 does not. Only the 10$\times$ LR arm moves any guard away from the group.}
  \label{tab:guards}
  \small
  \setlength{\tabcolsep}{5pt}
  \begin{tabular}{lcccccc}
    \toprule
    2.3B arm & $\Us$ & Deal rate & Deal-only $\Ut$ & Bad-parse (seller turns) & Turns & Buyer split \\
    \midrule
    base                                   & 0.422 & 0.939 & 0.469 & 0.0007 & 11.88 & $+0.051$ \\
    shared LR, 64 steps                    & 0.428 & 0.920 & 0.491 & 0.0013 & 12.61 & $+0.049$ \\
    shared LR, 104 steps                   & 0.424 & 0.909 & 0.496 & 0.0012 & 12.84 & $+0.079$ \\
    \textbf{3$\times$ LR, 56 steps}        & 0.460 & 0.948 & 0.501 & 0.0008 & 11.95 & $+0.032$ \\
    \textbf{10$\times$ LR, 56 steps}       & 0.473 & 0.876 & 0.582 & 0.0282 & 14.37 & $+0.175$ \\
    \bottomrule
  \end{tabular}
\end{table}%
}

\newcommand{\negscaleTableStrategy}{%
\begin{table}[t]
  \centering
  \caption{Post-hoc descriptors of the trained sellers' play (Appendix~\ref{app:deviations}). \emph{Opens with terms} is the share of episodes in which the seller names at least one of the domain's terms at turn 1, for the same-size base and then the arm. The $\Delta$ columns are paired differences from the same-size base over the scenarios both arms closed (969 to 1{,}152 of the 1{,}152 per arm), all exploratory; $\mathrm{e}$ marks a 95\% seed-cluster bootstrap interval that excludes zero (Table~\ref{tab:strategyfull}). Normalized utility splits exactly by term, into the continuous (distributive) and the categorical and binary (integrative) terms, with a term never named counted in its class, so up to rounding the first two $\Delta$ columns sum to the seller's deal-only change and the first three to the change in joint welfare.}
  \label{tab:strategy}
  \small
  \setlength{\tabcolsep}{5pt}
  \begin{tabular}{lccccc}
    \toprule
    & Opens with terms & \multicolumn{2}{c}{$\Delta$ seller $\Ut$} & & $\Delta$ joint \\
    \cmidrule(lr){3-4}
    Size & (\%, base $\to$ arm) & distributive & integrative & $\Delta$ buyer $\Ut$ & welfare \\
    \midrule
    \multicolumn{6}{l}{\emph{Shared LR, 104 steps}} \\
    2.3B  & 0.4 $\to$ 1.6 & $+0.010^{\mathrm{e}}$ & $+0.015^{\mathrm{e}}$ & $-0.013^{\mathrm{e}}$ & $+0.011$ \\
    4.5B  & 4.4 $\to$ 4.4 & $+0.006^{\mathrm{e}}$ & $+0.018^{\mathrm{e}}$ & $+0.000$ & $+0.024^{\mathrm{e}}$ \\
    12B   & 2.4 $\to$ 35.8 & $+0.049^{\mathrm{e}}$ & $+0.016^{\mathrm{e}}$ & $-0.057^{\mathrm{e}}$ & $+0.008$ \\
    31B   & 26.0 $\to$ 75.4 & $+0.062^{\mathrm{e}}$ & $+0.016^{\mathrm{e}}$ & $-0.072^{\mathrm{e}}$ & $+0.005$ \\
    \multicolumn{6}{l}{\emph{3$\times$ LR, step 56}} \\
    2.3B  & 0.4 $\to$ 0.1 & $+0.015^{\mathrm{e}}$ & $+0.019^{\mathrm{e}}$ & $-0.011^{\mathrm{e}}$ & $+0.023^{\mathrm{e}}$ \\
    4.5B  & 4.4 $\to$ 20.7 & $+0.037^{\mathrm{e}}$ & $+0.051^{\mathrm{e}}$ & $-0.055^{\mathrm{e}}$ & $+0.033^{\mathrm{e}}$ \\
    12B   & 2.4 $\to$ 76.6 & $+0.062^{\mathrm{e}}$ & $+0.015^{\mathrm{e}}$ & $-0.042^{\mathrm{e}}$ & $+0.035^{\mathrm{e}}$ \\
    31B   & 26.0 $\to$ 99.7 & $+0.065^{\mathrm{e}}$ & $+0.015^{\mathrm{e}}$ & $-0.039^{\mathrm{e}}$ & $+0.042^{\mathrm{e}}$ \\
    \multicolumn{6}{l}{\emph{10$\times$ LR, step 56}} \\
    2.3B  & 0.4 $\to$ 94.2 & $+0.061^{\mathrm{e}}$ & $+0.044^{\mathrm{e}}$ & $-0.105^{\mathrm{e}}$ & $+0.000$ \\
    \bottomrule
  \end{tabular}
\end{table}%
}

\newcommand{\negscaleTableStrategyFull}{%
\begin{table}[t]
  \centering
  \caption{Table~\ref{tab:strategy} with the 95\% seed-cluster bootstrap interval of every paired difference. Rows, columns and the $\mathrm{e}$ mark are as defined there.}
  \label{tab:strategyfull}
  \scriptsize
  \setlength{\tabcolsep}{3pt}
  \begin{tabular}{lccccc}
    \toprule
    & Opens with terms & \multicolumn{2}{c}{$\Delta$ seller $\Ut$} & & $\Delta$ joint \\
    \cmidrule(lr){3-4}
    Size & (\%, base $\to$ arm) & distributive & integrative & $\Delta$ buyer $\Ut$ & welfare \\
    \midrule
    \multicolumn{6}{l}{\emph{Shared LR, 104 steps}} \\
    2.3B  & 0.4 $\to$ 1.6 & $+0.010^{\mathrm{e}}$ \cis{+0.003}{+0.017} & $+0.015^{\mathrm{e}}$ \cis{+0.003}{+0.026} & $-0.013^{\mathrm{e}}$ \cis{-0.022}{-0.003} & $+0.011$ \cis{-0.003}{+0.026} \\
    4.5B  & 4.4 $\to$ 4.4 & $+0.006^{\mathrm{e}}$ \cis{+0.001}{+0.011} & $+0.018^{\mathrm{e}}$ \cis{+0.008}{+0.029} & $+0.000$ \cis{-0.009}{+0.010} & $+0.024^{\mathrm{e}}$ \cis{+0.012}{+0.037} \\
    12B   & 2.4 $\to$ 35.8 & $+0.049^{\mathrm{e}}$ \cis{+0.039}{+0.060} & $+0.016^{\mathrm{e}}$ \cis{+0.008}{+0.024} & $-0.057^{\mathrm{e}}$ \cis{-0.071}{-0.044} & $+0.008$ \cis{-0.008}{+0.023} \\
    31B   & 26.0 $\to$ 75.4 & $+0.062^{\mathrm{e}}$ \cis{+0.052}{+0.073} & $+0.016^{\mathrm{e}}$ \cis{+0.007}{+0.024} & $-0.072^{\mathrm{e}}$ \cis{-0.086}{-0.058} & $+0.005$ \cis{-0.013}{+0.025} \\
    \multicolumn{6}{l}{\emph{3$\times$ LR, step 56}} \\
    2.3B  & 0.4 $\to$ 0.1 & $+0.015^{\mathrm{e}}$ \cis{+0.009}{+0.023} & $+0.019^{\mathrm{e}}$ \cis{+0.004}{+0.033} & $-0.011^{\mathrm{e}}$ \cis{-0.022}{-0.000} & $+0.023^{\mathrm{e}}$ \cis{+0.004}{+0.042} \\
    4.5B  & 4.4 $\to$ 20.7 & $+0.037^{\mathrm{e}}$ \cis{+0.027}{+0.047} & $+0.051^{\mathrm{e}}$ \cis{+0.039}{+0.062} & $-0.055^{\mathrm{e}}$ \cis{-0.062}{-0.047} & $+0.033^{\mathrm{e}}$ \cis{+0.018}{+0.047} \\
    12B   & 2.4 $\to$ 76.6 & $+0.062^{\mathrm{e}}$ \cis{+0.051}{+0.073} & $+0.015^{\mathrm{e}}$ \cis{+0.003}{+0.026} & $-0.042^{\mathrm{e}}$ \cis{-0.063}{-0.022} & $+0.035^{\mathrm{e}}$ \cis{+0.012}{+0.057} \\
    31B   & 26.0 $\to$ 99.7 & $+0.065^{\mathrm{e}}$ \cis{+0.053}{+0.077} & $+0.015^{\mathrm{e}}$ \cis{+0.003}{+0.028} & $-0.039^{\mathrm{e}}$ \cis{-0.050}{-0.027} & $+0.042^{\mathrm{e}}$ \cis{+0.023}{+0.060} \\
    \multicolumn{6}{l}{\emph{10$\times$ LR, step 56}} \\
    2.3B  & 0.4 $\to$ 94.2 & $+0.061^{\mathrm{e}}$ \cis{+0.051}{+0.070} & $+0.044^{\mathrm{e}}$ \cis{+0.031}{+0.060} & $-0.105^{\mathrm{e}}$ \cis{-0.118}{-0.091} & $+0.000$ \cis{-0.018}{+0.022} \\
    \bottomrule
  \end{tabular}
\end{table}%
}

\title{Do Small Language Models Learn to Negotiate? \\
A Controlled Scaling Study of RL-Trained Sellers}

\author{%
  Pedro Tabacof \\
  Fin AI Research \\
  \texttt{pedro.tabacof@intercom.io} \\
  \And
  Sagar Joglekar \\
  Fin AI Research \\
  \texttt{sagar.joglekar@intercom.io} \\
}

\begin{document}

\maketitle

\begin{abstract}
LLM agents are starting to own the full customer experience. Soon, LLMs may be
selling and buying on behalf of companies and customers respectively. Small
models are more cost-efficient at scale, but can reinforcement learning train
them into competent sellers? We train four Gemma~4 checkpoints (2.3B to 31B
effective parameters) with GRPO on a programmatic utility reward for bilateral
multi-issue bargaining, and evaluate every arm on the same \pairedN{}
negotiations against two frontier buyers it never saw in training. With the
same learning rate ($10^{-6}$) for every size, the gain of the RL model over its
base rises from \gainAllEtwoPt{} at 2.3B to \gainAllThirtyOnePt{} at 31B. Each
size was trained once and the two smallest checkpoints use a different
architecture, so we fit no scaling law. Tripling the learning rate, with the
same or fewer training steps, improves on the shared rate at every size by
\tunedVsStepEtwoPt{} (2.3B) to \tunedVsStepEfourPt{} (4.5B). In exploratory
comparisons with two frontier models run as sellers, the 12B seller trained at
the tripled rate scores above both, though its untrained base already scores as
high as they do. The 4.5B seller at that rate shows no detectable difference
from either and fits on one 48\,GB GPU. A further 2.3B arm at
ten times the shared rate raises pooled score, but its gain concentrates on the
evaluation buyer that shares a model family with the training pool. These
results suggest tuning the learning rate before concluding that a small model
cannot learn to negotiate, and testing against buyers from more than one model
family.
\end{abstract}

\section{Introduction}
\label{sec:introduction}

Agents are beginning to act in workflows with direct economic consequences
(sales, procurement, marketplaces, contract negotiation), where negotiation
quality determines the value captured or conceded. Serving a frontier model for every recurring exchange
is costly and can conflict with latency, privacy and local-deployment
constraints, which favours small models specialized for agentic sub-tasks
\citep{belcak2025slmfuture}. Negotiation tests that specialization on a sub-task
where a weak agent concedes real value.
Prompting alone has not made LLMs strong negotiators. They model counterparty
preferences accurately yet fail to convert them into compensated trades
\citep{cosentino2026counterparty}, and only strong models improved from
self-play with in-context critique \citep{fu2023negotiation}. We therefore train
with a verifiable, outcome-aligned utility reward. In the closest prior work,
\citet{bergemann2026bilateral} fine-tune Qwen3 8B and 14B with GRPO for bilateral
trade but run the sizes as independent tracks ``with no cross-size comparisons''
and, with only two sizes, cannot ``separate capacity from pretraining
differences''. We
ask whether RL improves each size over its base, how the gain varies with size,
and whether that pattern changes with the learning rate.

\paragraph{Contributions.}
\textbf{(i) A shared-recipe comparison of the RL gain across four sizes on a
strategic, multi-agent task.} Under one GRPO recipe the paired gain over base
runs from \gainAllEtwoPt{} (not significant) at 2.3B to
\gainAllThirtyOnePt{} at 31B (Table~\ref{tab:factorial}). Each arm is a single
run, and size co-varies with architecture and training FLOPs. The gain's point
estimate rises by \diffEfourEtwoPt{} from 2.3B to 4.5B (interval includes zero), by
\diffTwelveEfourPt{} from 4.5B to 12B, where the architecture changes, and by
\diffThirtyOneTwelvePt{} from 12B to 31B (barely excludes zero), so about half the
rise coincides with the architecture change and we fit no scaling law.
\textbf{(ii) Evidence that the gain pattern across sizes depends on the learning
rate.} At the early checkpoints, tripling the rate improves on the shared-rate
arm at every size, with the largest point estimate at 4.5B (\tunedVsStepEfourPt{}).
Against a shared-rate run trained for 104 steps (56 for the tripled arm), the
tripled rate stays ahead at 2.3B and 4.5B, and the difference is not detectable
at 12B and 31B. Larger models score higher under both rates, and the base
models' point estimates already rank the same way.
\textbf{(iii) A demonstration that pooled score can conceal opponent-specific
gains.} A 2.3B arm at ten times the shared rate raises pooled score over base
while its gain
concentrates on the evaluation buyer that shares a model family with the
training pool, alongside formatting degeneration (Section~\ref{sec:results}).
\textbf{(iv) An evaluation protocol} of matched scenario cells, two buyers
excluded from training (one sharing the training pool's model family, one not),
a held-out domain, seed-cluster bootstrap intervals, and a deviation log.

\section{Related work}
\label{sec:related}

\paragraph{LLM negotiation and its evaluation.}
Bilateral multi-issue bargaining is the canonical setting in which distributive
and integrative issues coexist, so skill means both claiming a fixed pie and
enlarging it by logrolling \citep{walton1965behavioral, raiffa1982,
baarslag2013anac}. Recent benchmarks report surplus and diagnostic metrics
beyond deal rate, which conflates agreeableness with competence
\citep{zhang2026termsbench,
bianchi2024negotiationarena}. \citet{xia2024bargaining} report that larger models
are not necessarily better bargainers; they study \emph{base} models and we study
the gain from RL. Our held-out buyer panel follows the cross-play evaluation
of \citet{davidson2024agency} and the leaderboard-overfitting result of
\citet{singh2025leaderboardillusion}. Pre-LLM neural negotiators split a
dialogue-act strategy layer from generation \citep{lewis2017dealornodeal,
he2018decoupling}; our seller learns strategy inside the language
policy, with no act layer and no dialogue supervision. \citet{deepseekai2025r1}
warn that large-scale RL on a small base ``may not even achieve the performance
of distillation''; we test no distillation arm, and our 2.3B shared-rate
result is consistent with the warning. In adjacent settings,
\citet{liu2026spiral} train small models with zero-sum multi-turn self-play
where we use a fixed
external buyer pool; \citet{haque2026smallscale} study RL robustness for
small language-model agents.
\paragraph{RL for negotiation, and how the gain scales.} \citet{liu2026multibuyer} train
a 30B MoE model with RLVR in a multi-buyer market and conclude that RL on verifiable
environments ``is a more effective driver of utility than raw parameter scaling.''
That is a fixed-size comparison against larger untrained baselines. In our data,
RL improves on the base at 2.3B only once the learning rate is raised, and even
then the seller (\armEtwoTunedPt{}) scores below the untrained
12B base (\armTwelveBasePt{}). \citet{bergemann2026bilateral} is single-issue and distributive
where ours is multi-issue; its RL stage recovers deal rate while eroding surplus,
and at 2.3B our shared-rate arm does the opposite. On scaling, \citet{tan2026scalingbehaviors} fit RL
post-training on single-turn mathematical reasoning across the dense Qwen2.5
series and find larger models consistently more compute- and data-efficient, with
a saturation trend in the fitted form \citep{khatri2025artofscaling,
jin2025searchr1}; we know of no such study for a multi-agent task. In a learned-reward RLHF setting, \citet{hou2024rlhfscale} find that
``larger policy models benefit less
from RLHF with a fixed reward model.'' Size and compute allocation are not
separable, since larger policies buy fewer rollouts per FLOP
\citep{wilhelm2026rlcompute}.

\section{Environment, reward, and training}
\label{sec:method}

\begin{figure}[t]
  \centering
  \begin{tikzpicture}[
      font=\footnotesize,
      panel/.style={draw=black!55, rounded corners=2.5pt, align=left, inner sep=5pt, fill=white},
      actor/.style={panel, fill=black!4, minimum height=44pt},
      hero/.style={actor, draw=blue!55!black, fill=blue!6},
      steppanel/.style={panel, text width=104pt, minimum height=78pt},
      note/.style={font=\scriptsize, text=black!60},
      arr/.style={-{Latex[length=4.5pt, width=3.5pt]}, semithick, black!70},
      biarr/.style={{Latex[length=4.5pt, width=3.5pt]}-{Latex[length=4.5pt, width=3.5pt]}, semithick, black!70},
    ]

    \node[hero, text width=78pt, anchor=north west] (seller) at (0pt,0pt)
      {\textbf{Seller}\\Gemma~4,\\\textcolor{blue!45!black}{under training}};
    \node[actor, text width=136pt, anchor=north] (ledger) at (192pt,0pt)
      {\textbf{Offer ledger}~~\textcolor{black!55}{shared state}\\
       10 contract issues (car):\\
       ~~4 distributive (price, \ldots)\\
       ~~6 integrative (warranty, \ldots)};
    \node[actor, text width=78pt, anchor=north east] (buyer) at (384pt,0pt)
      {\textbf{Buyer}\\frontier model,\\frozen};
    \draw[biarr] ([yshift=-16pt]seller.north east) -- ([yshift=-16pt]ledger.north west);
    \draw[biarr] ([yshift=-16pt]ledger.north east) -- ([yshift=-16pt]buyer.north west);
    \node[note, anchor=north] at (192pt,-50pt)
      {each turn: two to three sentences of prose, then one JSON action};

    \coordinate (row2) at (0pt,-74pt);
    \node[steppanel, anchor=north west] (stepA) at (row2)
      {\textbf{1.~Seller turn}\\[2pt]
       \textsc{counter}, naming:\\
       \texttt{~~price: 40}\\
       \texttt{~~service: annual}\\
       \texttt{~~accessories: true}\\[2pt]
       \textcolor{black!60}{JSON action, parsed}};
    \node[steppanel, anchor=north] (stepB) at ([xshift=192pt]row2)
      {\textbf{2.~Ledger after the turn}\\[2pt]
       price: 40 (unchanged)\\
       service: annual (added)\\
       accessories: true (added)\\[2pt]
       \textcolor{black!60}{car model, interior carry over;\\unnamed terms add $0$ to $U$}};
    \node[steppanel, anchor=north east] (stepC) at ([xshift=384pt]row2)
      {\textbf{3.~Buyer turn}\\[2pt]
       \textsc{counter}: play returns\\to the seller\\[2pt]
       \textsc{accept}: episode ends\\[2pt]
       \textcolor{black!60}{no \textsc{accept} by round\\$T = 12$: no deal}};
    \draw[arr] (stepA.east) -- (stepB.west);
    \draw[arr] (stepB.east) -- (stepC.west);
    \node[draw=black!35, dashed, rounded corners=3pt, inner sep=5pt,
          fit=(stepA)(stepC), label={[note, anchor=south west]north west:one round}] (round) {};

    \node[panel, fill=black!4, text width=374pt, anchor=north west] (reward)
      at ([yshift=-14pt]round.south west -| stepA.west)
      {\textbf{Reward}, once per episode: $\Ut_{\mathrm{seller}}$ on a deal,
       $-\lambda = -0.3$ on a no-deal.~~
       \textcolor{black!60}{Equation~\ref{eq:reward} also subtracts small format and
       length penalties, omitted here.}};
    \draw[arr] (round.south -| stepC) -- (stepC |- reward.north)
      node[midway, right=2pt, note] {episode ends};
  \end{tikzpicture}
  \caption{The negotiation loop. Panels 1--2 show the seller turn quoted in the
  text and the ledger after it, with the JSON action as parsed terms. The buyer
  sees the seller's prose and the resulting ledger, and utility is scored on the
  ledger at the end. Evaluation buyers are the held-out pair of
  Section~\ref{sec:setup}; the training buyers were GPT-5.5 and GPT-5.4-mini.}
  \label{fig:setup}
\end{figure}
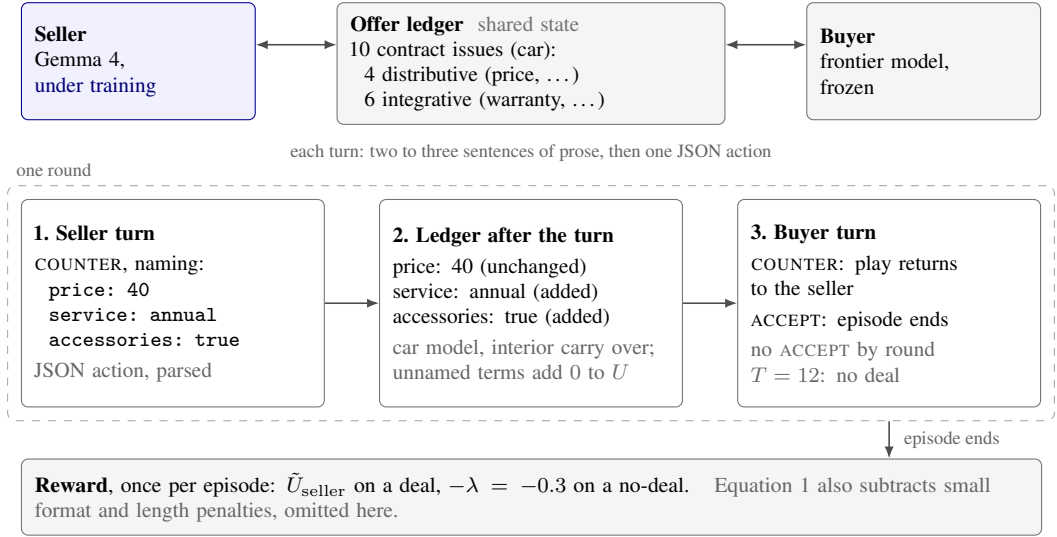

\paragraph{Environment and action protocol.}
A scenario is a bilateral negotiation between a \emph{seller} (the trained policy)
and a \emph{buyer} (a frozen LLM) over a contract with $7$ to $10$ issues ($10$ in
the car domain): $3$ to $5$ distributive continuous terms whose preference
direction is fixed and opposed by role, and integrative terms on which each party
independently draws a preferred level, so
differences in private weights can create joint gains through logrolling. Utility is linear in
a feature map, $U(x) = \sum_k w_k \phi_k(x)$, with importances drawn per party from a
$\mathrm{Dirichlet}(0.3)$ prior, and we report $\Ut = (U(x) - U_{\mathrm{min}}) /
(U_{\mathrm{max}} - U_{\mathrm{min}}) \in [0,1]$, where $U_{\mathrm{min}}$ and
$U_{\mathrm{max}}$ are that party's worst and best feasible
contracts. Each party sees its own weights and nothing of the other's.
The seller opens and the parties alternate; a round is one seller turn and one
buyer turn, for at most $T = 12$ rounds ($24$ turns). A turn is two to three
sentences of prose followed by one JSON action, \textsc{counter} or
\textsc{accept}. Terms named in a \textsc{counter} overwrite a persistent ledger
that holds every term named so far, and the buyer reads the prose and the
resulting ledger but not the JSON. A term never named contributes $0$ to $U$;
because $0$ can lie above that term's worst level, an unnamed term can still add
to $\Ut$. In evaluation, terms attached to an
\textsc{accept} are ignored and the episode closes on the current ledger;
reaching the cap without one is a no-deal. The buyer is a frozen prompted model
with no reward. Each party's prompt says any deal above its reservation value
beats no deal, and neither states the 24-turn cap (evaluation prompts show the
turn number). Below are turns 6 and 7 of one evaluation episode on the held-out car
domain, between GPT-5.6-Terra and the 12B seller trained at a learning rate of
$3{\times}10^{-6}$, verbatim except that the action's free-text notes field is
omitted (the action gives prices in \$k). Figure~\ref{fig:setup} shows the action
as parsed terms, and Appendix~\ref{app:example} describes how the episode closes.

\smallskip
\noindent\fbox{\parbox{\dimexpr\linewidth-2\fboxsep-2\fboxrule\relax}{\small\microtypesetup{protrusion=false}
\textbf{Buyer (turn 6).} ``I can accommodate the Premium interior, but \$42,500 is
still above my budget. If we keep the Sedan with Premium interior, I can move
forward at \$40,000.''\\[3pt]
\textbf{Seller (turn 7).} ``I really appreciate you being willing to keep the
Premium interior; that means a lot to me. While \$40,000 is a bit lower than I
hoped, I'd be willing to accept that if we can include an annual service plan and
accessories in the deal.''\\[3pt]
\frenchspacing\texttt{\{"action": "COUNTER", "terms": [\{"name": "price", "value": 40\},}\\
\texttt{\phantom{\{"action": "COUNTER", "terms": [}\{"name": "service", "value": "annual"\},}\\
\texttt{\phantom{\{"action": "COUNTER", "terms": [}\{"name": "accessories", "value": true\}]\}}}}
\smallskip

\paragraph{Reward.}
The reward is terminal and programmatic, with no per-turn shaping and no model
in the loop: the seller is paid its own normalized utility when a deal closes,
a fixed penalty when none does, minus small penalties for malformed turns and
for dragging the episode out. With $\lambda = 0.3$, $g_{\mathrm{fmt}} =
g_{\mathrm{len}} = 0.05$ and soft budget $t_{\mathrm{soft}} = 0.75\,T$,
\begin{equation}
  r \;=\;
  \underbrace{\Big[\Ut\ \text{on agreement},\ \ -\lambda\ \text{on no-deal}\Big]}_{\text{outcome}}
  \;-\; g_{\mathrm{fmt}}\,\frac{n_{\mathrm{bad}}}{n_{\mathrm{policy}}}
  \;-\; g_{\mathrm{len}}\,\frac{\max(0,\, n - t_{\mathrm{soft}})}{T - t_{\mathrm{soft}}},
  \label{eq:reward}
\end{equation}
where $n_{\mathrm{bad}}/n_{\mathrm{policy}}$ is the fraction of the seller's own
turns that failed to parse and $n$ is the episode's realized length in rounds.
The constants were fixed once, before any training, are identical at every
size, and were never tuned. The no-deal penalty
sits strictly below $U_{\mathrm{min}}$, so the seller has no outside option
(reaching the horizon without agreement earns less than accepting even its
worst feasible contract), and the two penalties are several times smaller
than the outcome term. We avoid per-turn shaping because it needs a judge or a
concession heuristic that bakes in the strategy prior the policy should learn
\citep{wei2025turnlevel}.

\paragraph{Optimization.} We train with NeMo-RL's GRPO-style estimator
\citep{shao2024deepseekmath, nemorl} (group-normalized advantages with a
leave-one-out baseline, no learned value model) on its multi-turn rollout
driver: each seller turn is one generation call, the buyer's reply
is the observation, and a dialogue is one trajectory with the buyer's tokens masked
out of the loss. Apart from the learning rate and the evaluated checkpoint, every
arm runs the same recipe: $128$ episodes per step, a warmup--cosine schedule, KL
penalty $0.04$ to the initial policy, one training run on a single seed per
size-by-rate arm (Appendix~\ref{app:recipe}; only tensor-parallel degree and
generation memory fraction differ by size). The shared rate, $10^{-6}$, was not tuned per
size; Section~\ref{sec:setup} adds higher rates.

\section{Experimental setup}
\label{sec:setup}

\paragraph{Sizes, architectures, and the step budget.}
Four Gemma~4 sellers span $13\times$ in effective parameters and $6\times$ in total
parameters \citep{gemma4}. All four are dense. The two \textbf{E-series}
checkpoints use per-layer embeddings and have fewer effective than stored
parameters (2.3B of 5.1B and 4.5B of 8.0B); \textbf{12B and 31B} are
conventional checkpoints. Architecture co-varies with size (Gemma~4 ships no
conventional checkpoint below 12B),
so no cross-size contrast isolates parameter count. Each size is evaluated as a base
instruction-tuned seller and after GRPO under an identical inference configuration,
so the only configured difference inside a base-versus-RL contrast is the
weights. Arms are matched on \textbf{optimizer steps}. The
larger arms therefore use more FLOPs, and every cross-size contrast confounds
size with compute (Section~\ref{sec:discussion}).

\paragraph{Learning-rate arms.} Arms are named by their multiple of the shared
value. Every size gets 104 combined steps at the \emph{shared LR} ($10^{-6}$):
an initial run plus a continuation under a fresh warmup--cosine schedule. A
separate 64-step arm at the \emph{3$\times$~LR} ($3{\times}10^{-6}$) is trained
at each size with everything else unchanged and evaluated (``read'') at step
56: step-matched to the shared-LR initial runs at 4.5B and 31B, and eight steps
short of them at 2.3B and 12B (per-size schedules and one storage fault:
Appendix~\ref{app:deviations}). One further 2.3B arm runs at the
\emph{10$\times$~LR} ($10^{-5}$).

\paragraph{Held-out buyers, and a held-out domain.} Training buyers are GPT-5.5
(20\%) and GPT-5.4-mini (80\%), attached to each
scenario as a dataset field, so buyer mixture and scenario assignment are
identical across arms.
Evaluation buyers are Claude Sonnet~5 and GPT-5.6-Terra, held out by \emph{model}
(neither is in the training pool) but not by model family, since Terra is a
later GPT model. A gain concentrated on the same-family buyer would suggest
a policy that does not generalize across counterparties, though with one model
per family we cannot separate a family effect from a model-specific one.
Training spans eight commercial domains; a ninth, a used-car
sale, is withheld (a new surface domain over the same issue structure). Both
frontier models are also run \emph{as sellers} under the identical protocol,
though their decoding differs from the self-served arms'
(Appendix~\ref{app:limitations}).

\paragraph{Paired evaluation, metrics, and guards.}
Every seller arm faces the identical scenario set: the same $64$ seeds per domain, the
same private weight draws, both buyers, nine domains, \pairedN{} episodes per arm
(of which \heldoutN{} are on the held-out domain), \totalEpisodes{} episodes over
\numArms{} arms, with evaluation seeds and weight draws disjoint from training.
Contrasts are within-scenario paired
differences; intervals are percentile bootstraps with $10{,}000$ resamples in which
\emph{seeds are resampled as clusters}, so a resampled seed carries all its domains
and both buyers \citep{miller2024errorbars, bowyer2025noclt}. The effective sample
size is \textbf{64 seed clusters} of 18 episodes, not \pairedN{} independent
episodes. We score $\Us$, the reward's outcome term ($\Us = \Ut \in [0,1]$ on a
deal and $-0.3$ on a no-deal), so $\Us \in [-0.3, 1]$ and no-deal episodes
contribute instead of being dropped. Alongside $\Us$ we report deal rate, deal-only $\Ut$, joint
welfare (the sum of both parties' utilities), mean
turns and parse-failure rate \citep{zhang2026termsbench}, since deal-only
utility can rise while overall score falls. The four prespecified guards are a
deal rate that holds while $\Ut$ rises, a flat parse-failure rate, mean turns
below the cap, and hand-read transcripts from every trained arm
(Appendix~\ref{app:degenerate}).

\paragraph{Two significance regimes.} The primary family is the shared-LR gain
over base at each size (four prespecified contrasts), Holm-corrected at
$\alpha = 0.05$. The 4.5B, 12B and 31B gains are significant (marked $*$) and the
2.3B gain is not, with or without correction (p-values in
Appendix~\ref{app:deviations}). All other intervals, including every 3$\times$
LR contrast and every comparison with a frontier seller, are exploratory and
unadjusted (marked $\mathrm{e}$ where they exclude zero). Every interval is
pointwise. Evaluation checkpoints are fixed by the step-matching rule and were not
chosen on validation; some arms happen to read at their peak
(Section~\ref{sec:results}).
Appendix~\ref{app:deviations} logs every departure from the frozen protocol
\citep{baker2025monitoring}, including a prespecified doubling of the held-out
domain's episodes that triggered but was not run because of cost, so held-out
analyses may be underpowered (Appendix~\ref{app:heldout}).

\section{Results}
\label{sec:results}

\begin{figure}[t]
  \centering
  \includegraphics[width=\linewidth]{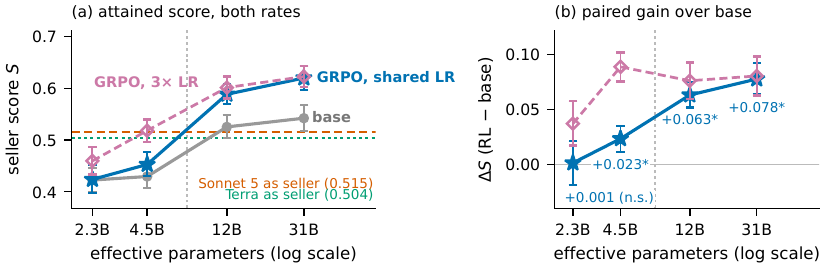}
  \caption{(a) Attained seller score $\Us$ (\emph{arm-level} means, \pairedN{} paired
  episodes per arm, nine domains, two held-out frontier buyers) for base, the shared
  LR ($10^{-6}$, 104 steps) and the 3$\times$~LR (step 56). Horizontal lines
  are the frontier reference sellers, connecting lines are visual guides only,
  and the vertical dotted line marks the E-series/conventional architecture boundary
  (Section~\ref{sec:setup}). (b) The trained arms as \emph{paired} differences
  from base. Error bars are 95\% intervals on within-scenario paired
  differences and are not comparable with (a)'s arm-level intervals; point
  labels give the shared-LR gains ($*$ survives Holm correction, n.s.\ does
  not). Every other interval is exploratory and unadjusted; contrasts between
  the two rates are in Table~\ref{tab:factorial}.}
  \label{fig:scaling}
\end{figure}

\negscaleTableFactorial

\paragraph{At 104 steps under the shared LR, the gain point estimates rise with size.}
The paired gain over base (Figure~\ref{fig:scaling}, Table~\ref{tab:factorial})
is \gainAllEtwo{} at 2.3B and \gainAllEfour{} at
4.5B against \gainAllTwelve{} at 12B and
\gainAllThirtyOne{} at 31B, and at the early checkpoints the point estimates
of attained score also rise with size (Table~\ref{tab:grid}). These are score units on the
$[-0.3, 1]$ scale of Section~\ref{sec:setup}, not a percentage of the
bargaining range. The intervals on the four primary contrasts capture scenario resampling only.
Each arm was trained once, so there is no p-value for the size trend and no
estimate of run-to-run variance. As a rough stability check, the largest movement in $\Us$ between adjacent
read points runs from $0.004$ at 2.3B to $0.056$ at 31B (Table~\ref{tab:grid}),
and at 31B it spans 48 further steps after a schedule restart. Within a single
run, the 4.5B 3$\times$~LR arm's read points at steps 56 and 64 differ by
\efourStepCheck{}. The two low-gain
arms are the E-series checkpoints and the two high-gain arms are conventional, so
the design does not separate size from architecture. The gain \emph{difference}
is \diffEfourEtwo{} within the E-series pair, \diffTwelveEfour{} across the
boundary and \diffThirtyOneTwelve{} within the conventional pair (unadjusted).

\paragraph{The 3$\times$~LR helps at every size at the early checkpoints.}
With no per-size tuning, the small gains at 2.3B and 4.5B might reflect the
learning rate rather than size. Against base the 3$\times$~LR arm gains at every
size, with the largest point estimate at 4.5B (\tunedAllEfour{}). Against the
shared-LR initial run at the early checkpoints it also gains at every size
(Table~\ref{tab:factorial}, column 5).

\paragraph{With 104 steps, the shared LR is not detectably behind at 12B and 31B.}
The 104-step shared-LR arm trains for roughly twice the 3$\times$~LR arm's 56
steps, so the two are not compute-matched; an operator choosing between a higher
rate and longer training faces this comparison. Against the 104-step arm the
3$\times$~LR arm stays ahead at 2.3B (\tunedVsContEtwo{}) and
4.5B (\tunedVsContEfour{}), while at 12B and 31B the intervals include zero
(Table~\ref{tab:factorial}, column 6). The ordering of mean attained scores is
the same under both rates: the 3$\times$~LR arms score \armEtwoTunedPt{} $<$
\armEfourTunedPt{} $<$ \armTwelveTunedPt{} $<$ \armThirtyOneTunedPt{}. That
12B-to-31B difference barely excludes zero (\tunedStepThirtyOneTwelve{},
unadjusted) and is confounded by read point, since 12B is read 24 steps after its
validation peak and 31B at its own. At 31B the 3$\times$~LR arm is not
detectably different from the 104-step arm, but \openerTunedThirtyOne{} of its
seller turns begin with the same three words, against \openerIncThirtyOne{} for
that arm (a post-hoc diagnostic; Appendix~\ref{app:degenerate}).

\paragraph{Frontier reference sellers (exploratory).}
In unadjusted contrasts, the 3$\times$~LR
12B seller scores above Claude Sonnet~5 by \tunedVsSonnetTwelve{} and above
GPT-5.6-Terra by \tunedVsTerraTwelve{}, though the untrained 12B base
(\armTwelveBasePt{}) already scores as high as both (\armSonnetPt{} and
\armTerraPt{}). The 3$\times$~LR 4.5B seller shows no detectable difference from
either (\tunedVsSonnetEfour{}, \tunedVsTerraEfour{}; not an equivalence test) and
fits on one 48\,GB accelerator. Decoding differs by provider and the reference
arms include self-play (Appendix~\ref{app:limitations}); Appendix~\ref{app:efficiency}
covers serving.

\paragraph{The 10$\times$~LR arm's pooled gain concentrates on the same-family buyer.}
At 2.3B, where the 3$\times$~LR arm passes every guard, the 10$\times$~LR arm
scores \pushedGain{} over base, but its buyer split ($\Us$ against Terra minus
$\Us$ against Sonnet~5) is \pushedBuyerGap{}, compared with \splitRangeLo{} to
\splitRangeHi{} for the other 2.3B arms. Its format also degenerates,
with \jsonLeak{} of its seller turns leaking a fragment of the JSON action into the
prose the buyer reads and 53.4\% beginning with ``I appreciate'' (17.5\% for the
3$\times$~LR arm; Appendix~\ref{app:degenerate}).

\paragraph{The size pattern is weakest on the held-out domain.}
On the withheld car-sale domain ($n = \heldoutN{}$) estimates are imprecise, and
only two of the eight gains exclude zero, those of the 104-step 12B arm
(\gainCarTwelve{}) and the 3$\times$~LR 2.3B arm (\tunedCarEtwo{};
Appendix~\ref{app:heldout}). The 12B and 31B arms still score highest there,
but the 2.3B arms score above the 4.5B arms under both rates, so the full
ordering does not replicate and we do not claim the \emph{gain} pattern does.

\negscaleTableStrategy

\paragraph{What the trained sellers do differently (post hoc).}
We chose the descriptors in Table~\ref{tab:strategy} after the results were
known (Appendix~\ref{app:deviations}). The seller always opens, so only its prompt
carries the turn-1 instruction to start conversationally and to name only the
terms its dialogue mentions. The base sellers name at least one term at turn 1
in \stratOpenLo{} to \stratOpenHi{} of episodes. The 3$\times$~LR sellers at 12B and
31B do so in \stratOpenTwelveTuned{} and \stratOpenThirtyOneTuned{} of episodes,
naming \stratTermsTwelveTuned{} and \stratTermsThirtyOneTuned{} terms on average
(\stratTermShareTwelveTuned{} and \stratTermShareThirtyOneTuned{} of the domain's).
Accepted as they stand, those openings would score \stratOpenUTwelveTuned{} and
\stratOpenUThirtyOneTuned{} for the seller, and the deals they lead to score
\stratConcTwelveTuned{} and \stratConcThirtyOneTuned{} lower. Agreements between
prompted LLM negotiators track the opening anchor \citep{cosentino2026counterparty},
and LLM buyers favour the first proposal they receive \citep{bansal2025magentic}.
These openings are consistent with both. We cannot tell from
Table~\ref{tab:strategy} whether the opening causes the seller's gain, and the
4.5B 3$\times$~LR seller, with the largest pooled gain over base
(\tunedAllEfourPt{}), opens with terms in only \stratOpenEfourTuned{} of episodes.

At 12B and 31B, under both rates, the distributive terms account for
\stratDistShareLo{} to \stratDistShareHi{} of the seller's gain on the scenarios both
arms closed. Under the shared LR most of that gain comes out of the buyer's share.
The buyer's loss is \stratBuyerShareTwelveInc{} and \stratBuyerShareThirtyOneInc{}
of it, and joint welfare does not move detectably (\stratWelfTwelveInc{} and
\stratWelfThirtyOneInc{}). Under the 3$\times$~LR
the buyer's loss is \stratBuyerShareTwelveTuned{} and
\stratBuyerShareThirtyOneTuned{} of the gain, and welfare rises
(\stratWelfTwelveTuned{} and \stratWelfThirtyOneTuned{}). An integrative term
can still move value between the parties. At 31B under the shared LR the buyer's
change on those terms is \stratBuyerIntegThirtyOneInc{} and the seller's
\stratIntegThirtyOneInc{}. At
2.3B and 4.5B the integrative part has the larger point estimate under both rates,
but the gap excludes zero only at 4.5B under the shared LR (\stratGapEfourInc{}).
From turn 16 each party's prompt says to converge,
and \stratLateBaseLo{} to \stratLateBaseHi{} of base deals close at or after that
turn. Training changes that share by \stratLateLo{} to \stratLateHi{} points on
scenarios both arms closed, and by \stratLatePushed{} for the 10$\times$~LR arm.

\section{Discussion}
\label{sec:discussion}

\paragraph{A programmatic reward is still a proxy.}
Writing the reward in code removes the learned reward model's approximation
error \citep{hou2024rlhfscale, gao2022overoptimization}, but the
policy is still scored against a frozen \emph{opponent pool} that defines the
training objective; pushed hard enough, it exploits the pool: in the
10$\times$~LR arm's no-deal episodes the seller counters without conceding, and
the \pushedBuyerGap{} buyer split shows the pooled gain concentrating on the
same-family buyer (Appendix~\ref{app:degenerate}). In our runs the logged KL to
base did not track guard failures. The 31B 3$\times$~LR arm has the largest
step-56 KL (\klThirtyOneTuned{}) and passes all four prespecified guards, while
the 10$\times$~LR arm, at a KL of \klPushed{}, fails three (Appendix~\ref{app:diagnostics}).

\paragraph{Compute, tuning and serving.}
At 104 steps the 31B arm used $6$--$13\times$ the 2.3B arm's training FLOPs (Appendix~\ref{app:flops}), so the shared-LR pattern could also be a compute
artefact \citep{wilhelm2026rlcompute}. The 3$\times$~LR arms test the
learning-rate explanation only in part (two rates, no per-size tuning). At 2.3B
and 4.5B, raising the rate improved the score more than extending the shared-LR run to 104 steps,
but larger sizes still receive more FLOPs at matched steps, so the compute
explanation stays open. At serving time, the 12B 3$\times$~LR
seller scores \armTwelveTunedPt{} on one 48\,GB device against the 31B base's
\armThirtyOneBasePt{} on four (we do not benchmark throughput, latency or
cost; Appendix~\ref{app:efficiency}).

\textbf{Limitations.} We use one environment class (no outside option; the
seller always opens), one model family and RL algorithm, a seller-side objective,
one training run per arm, two learning rates and no scripted baseline seller. Arms
are not compute-matched, read points are unequal at 2.3B and 12B, size and
architecture co-vary between 4.5B and 12B, and reference sellers decode
differently. Evaluation covers two buyers (one per model family) and one held-out
domain ($n = \heldoutN{}$). All intervals condition on the nine domains, two buyers
and realized runs. The strategy descriptors of Table~\ref{tab:strategy} are
post hoc (Appendix~\ref{app:limitations}).

\paragraph{Conclusion.}
Across four Gemma~4 checkpoints trained once each, the shared-LR gain point
estimates rise with size, and at the early checkpoints tripling the learning
rate improves on the shared rate at every size, with the largest point estimate
at 4.5B. Larger checkpoints score higher under both rates, and the bases' point
estimates already rank the same way. The 10$\times$~LR arm raises pooled score,
but its gain concentrates on the same-family buyer. These results suggest two
practical lessons: tune the learning rate before concluding that a small model
cannot learn to negotiate, and evaluate it against buyers from more than one
model family.

\begin{ack}
We thank Fergal Reid for his review of the paper and his suggestions, and the
anonymous reviewers for their feedback. This work received no third-party
funding or support, and the authors have no competing interests to declare.
\end{ack}

\bibliographystyle{plainnat}
\bibliography{refs}

\appendix

\section{Limitations in full}
\label{app:limitations}

\textbf{One environment class.} The negotiation is bilateral and multi-issue,
with private per-party weights, no information-asymmetry condition, 12 rounds and
one issue schema. The seller always opens, so turn-order effects are untested.
Single-issue distributive bargaining \citep{bergemann2026bilateral} and
multi-party settings may behave differently.

\textbf{Size and architecture co-vary.} The two
low-gain arms are E-series checkpoints (per-layer embeddings) and the two high-gain arms are
conventional, so the large step in the gain at the 4.5B-to-12B boundary is equally
consistent with ``conventional Gemma~4 checkpoints respond to GRPO and the
E-series does not.'' Gemma~4 publishes no conventional checkpoint below 12B and no
E-series checkpoint above 4.5B, so within this family the two cannot be separated,
and separating them would need a second family with conventional small models. The
gain's point estimate rises \emph{within} both pairs, by \diffThirtyOneTwelve{}
in the conventional pair (only marginally distinguishable from zero at $10^{5}$
resamples, unadjusted) and by \diffEfourEtwo{} in the E-series pair (interval
straddling zero). An architecture-only account predicts no rise within a pair,
but neither rise is well resolved. Across the boundary, where architecture
changes with size, the difference is \diffTwelveEfour{}.

\textbf{One model family, one seed
per size.} Varying parameter count within one Gemma~4 release buys the clean
comparison and forbids any cross-family claim; RLVR effects are known to be
family-specific \citep{shao2025spuriousrewards}. Each size was trained once, so
we cannot separate the size trend from run-to-run training variance, and the
intervals quantify scenario sampling only.

\textbf{Matched steps, not matched FLOPs.} Section~\ref{sec:discussion} discusses
this, and Appendix~\ref{app:flops} gives the FLOP accounting.

\textbf{Seller-side only.} The buyer side and co-adaptation between the two are
untested, and frozen opponents measure competence against a fixed distribution
rather than equilibrium behaviour.

\textbf{Two learning rates, not a sweep.} One rate is shared across all four
sizes, with no per-size tuning, and the factorial adds one alternative rate at
every size rather than an optimum at any of them. At the early checkpoints
(equal or fewer steps) the 3$\times$~LR improved on the shared rate at every size
(Section~\ref{sec:results}). After 104 steps the shared rate gains
\gainAllEtwoPt{} over base at 2.3B and \gainAllEfourPt{} at 4.5B, against
\tunedAllEtwoPt{} and \tunedAllEfourPt{} for the 3$\times$~LR at step 56. At 12B
and 31B the two rates are then no longer detectably different. A rate between or
beyond the two we ran could move any point, and each 3$\times$~LR arm is read at
one step. Establishing the curve under per-size optima needs a full sweep at four
sizes, several times the training budget this study spent.

\textbf{One RL algorithm, one reward shape.} Turn-level shaping
\citep{wei2025turnlevel} or hierarchical credit assignment \citep{zhou2024archer}
might change the small-model result again.

\textbf{Two evaluation buyers.} They are held out by model, one in the training
pool's provider family and one outside it, but there are only two, and LLM
strategic behaviour is opponent-conditional \citep{akata2025repeatedgames}.

\textbf{One held-out domain at $n = \heldoutN{}$.} The panel is underpowered, and
the prespecified extension was not run (Appendix~\ref{app:heldout}).

\textbf{Partial strategy analysis.} Table~\ref{tab:strategy} reports how often
each arm names terms at turn 1 and splits its change in utility into distributive
and integrative parts, using descriptors chosen after the results were known.
Testing whether the opening causes the gains would need sellers held to a fixed
opening. We did not compute Nash distance or Pareto efficiency, so joint welfare
is the only efficiency proxy we report.

\textbf{No scripted baseline.} Every reference seller is a model (the four bases
and two frontier sellers). A scripted seller, for example one that names a fixed
offer and waits for the buyer to accept it, would show how much of each arm's
score a fixed strategy already reaches against these buyers; we did not run one.

\textbf{Off-schema terms.} A term name outside a domain's schema (for example
\texttt{delivery\_day} for the car domain's \texttt{delivery}) enters the ledger
but is never scored, and the schema term it stands for counts as never named
unless the ledger also holds it. Such names are 4--6\% of the terms sellers name
in every arm. The share of deals that close with at least one on the ledger is
55\% for the 10$\times$~LR arm and 27--44\% for every other arm, highest for the
12B and 31B 3$\times$~LR arms (44\% and 43\%, against 27--32\% for the other 12B
and 31B arms).

\textbf{Training and evaluation differ in two ways.} In training, the seller's
context after its first turn grows by the buyer's prose alone, while the evaluation
harness re-renders the full scaffold (history, current offer, phase instruction
and format reminder) on every seller turn. Every arm is evaluated with that
scaffold, but only the trained arms were optimized on the training prompt, so
their gains are measured under a prompt shift. For the same reason, absolute
$\Us$ here is not directly comparable to a training-validation reward. Training
also applied terms the seller attached to an \textsc{accept} before closing the
deal, so a seller could close on terms the buyer never offered; evaluation
ignores them. In the evaluation transcripts such \textsc{accept}s are 0--5\% of
the seller's closing \textsc{accept}s in the base and shared-LR arms, 5--10\% in
the 3$\times$~LR arms above 2.3B and 56\% in the 10$\times$~LR arm.

\textbf{Sampling, exactly as sent.} Every self-served Gemma seller, base and RL
alike, runs the training rollout's own settings (temperature $1.0$, top-$p$
$0.95$, top-$k$ $64$), and the harness warns at startup if they drift. The two frontier
reference \emph{sellers} were requested at temperature $0.7$ instead; Terra
honours that value and Claude Sonnet~5 rejects the parameter and runs at its
provider default, so the Sonnet~5 seller is a reference point only.
Both buyers run at $1.0$, Terra by request and Sonnet~5 by provider default (it
rejects the parameter).

\textbf{Self-play in the reference arms.} Each frontier seller meets a same-model
buyer on \sonnetSelfPlayN{} of its \pairedN{} episodes, which no Gemma arm does.
As a seller Sonnet~5 scores \sonnetVsSonnetBuyer{} against itself and
\sonnetVsTerraBuyer{} against the other buyer; Terra \terraVsTerraBuyer{} and
\terraVsSonnetBuyer{}. Dropping the self-play half would barely move either
reference score.

\section{Full evaluation grid and the 2.3B deal-rate decomposition}
\label{app:grid}

\begin{figure}[h]
  \centering
  \includegraphics[width=\linewidth]{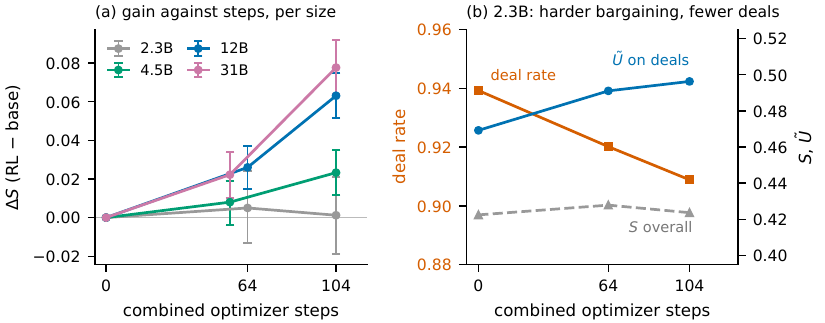}
  \caption{(a) Paired gain over base against combined optimizer steps, per
  size, with 95\% intervals. The intermediate point is step 64 at 2.3B and 12B and
  step 56 at 4.5B and 31B, so the continuation adds 40 steps at the first two and
  48 at the second two. Point estimates of the gain from those added steps rise
  with size (\contAllEtwoPt{}, \contAllEfourPt{}, \contAllTwelvePt{},
  \contAllThirtyOnePt{}).
  (b) What the shared recipe does at 2.3B: deal rate falls from 0.939 to 0.909
  while utility per closed deal rises from 0.469 to 0.496, leaving the score
  flat. Table~\ref{tab:guards} shows a larger learning rate moves both.}
  \label{fig:mechanism}
\end{figure}

Table~\ref{tab:main} decomposes the headline arms into deal rate, utility per
closed deal and joint welfare; Table~\ref{tab:grid} is the full grid behind
Figure~\ref{fig:scaling}, including the intermediate checkpoints and both
learning rates. Every arm is evaluated on the identical \pairedN{} paired
episodes.
Figure~\ref{fig:mechanism} is the gain-versus-steps view of the shared-recipe
arms discussed in Section~\ref{sec:results}.

\negscaleTableMain

\negscaleTableGrid

\section{What the held-out panel separates}
\label{app:heldout}

The four shared-recipe gains on the withheld car-sale domain are
\gainCarEtwo{} at 2.3B, \gainCarEfour{} at 4.5B, \gainCarTwelve{} at 12B and
\gainCarThirtyOne{} at 31B. Under the 3$\times$~LR they are \tunedCarEtwo{},
\tunedCarEfour{}, \tunedCarTwelve{} and \tunedCarThirtyOne{}. Two of the eight
intervals exclude zero, the 104-step 12B arm's and the 2.3B 3$\times$~LR
arm's, and only the 2.3B interval clears zero by more than the Monte Carlo band.
The 31B 3$\times$~LR arm, which has the highest pooled score of any arm, has the
smallest held-out gain of the four 3$\times$~LR arms.
In its car transcripts it opens in the buyer's voice on \carVoiceTuned{} of
episodes, against \carVoiceInc{} for the 104-step 31B arm on the same detector.

Two further findings rest on the held-out panel
rather than the pooled score. First, the two frontier sellers are indistinguishable on the
training domains (\frontierTrain{}, Sonnet~5 minus Terra) and pooled
(\frontierAll{}), while held out Sonnet~5 leads by \frontierCar{}. Second, the
panel reorders the trained arms. With 31B read at step 56 against the 104-step
12B arm, 31B lost
pooled (\earlyFlipAll{}) but won held out (\earlyFlipCar{}). At matched
104 steps the flip disappears and 31B wins both panels, which is consistent with a
compute artefact, though one reading point cannot show that it is one. Both are exploratory contrasts at
$n = \heldoutN{}$.

The protocol prespecified one adaptive step, doubling the held-out panel if its
interval straddled zero for the largest trained model. That condition \emph{is}
met, and now twice (\gainCarThirtyOne{} for the 104-step 31B arm and
\tunedCarThirtyOne{} for the 3$\times$~LR arm), and we did not run the extension. The
protocol requires it be applied at once to both buyers and to all ten seller arms of
the frozen arm matrix (four bases, four shared-recipe arms, two frontier sellers; the
\numArms{} arms evaluated here add intermediate checkpoints and the 3$\times$ and
10$\times$~LR arms to those), and then never re-taken, which is outside the budget of either
evaluation round that met the condition. It is recorded as a live follow-up in the deviation log rather than
taken unilaterally. A held-out interval that includes zero therefore does
not show that the effect is absent. At $n = \heldoutN{}$ a 95\% interval has a
half-width of about $\pm0.03$, which is the size of the effects being tested.

\section{Training diagnostics: the small arm was moved least}
\label{app:diagnostics}

The 2.3B deal-rate decline in Figure~\ref{fig:mechanism}b is concentrated
in \texttt{acquisition}, \texttt{job\_offer} and
\texttt{partnership\_dissolution}, where its shared-recipe deal rates fall to
$0.74$--$0.80$; five of the other six stay above $0.93$ and
\texttt{media\_rights} sits at $0.90$.

A flat gain has two readings. Either the policy collapsed onto a narrow
strategy and stopped exploring, or the optimizer never moved it. The training
logs support the second, and the 3$\times$~LR arm is consistent with it. Over the
continued steps the policy entropy is flat at 2.3B ($0.269 \to 0.261$) and
higher than either conventional arm's throughout (12B
$0.139 \to 0.126$; 31B $0.099 \to 0.111$, the only arm whose entropy rises);
4.5B, at $0.466 \to 0.443$, is higher still.
All KL figures below are the logged \texttt{train/kl\_penalty} series, the
$k_3$ estimate of KL to the initial policy; the $0.04$ coefficient, identical
at every size, is applied in the loss and not folded into the logged value.
At the \emph{end} of each shared-recipe continuation the KL
orders the other way from entropy and grows with size:
$0.0088$ at 2.3B, $0.0157$ at 4.5B, $0.022$ at 12B and $0.0485$ at 31B. The small
policy stayed the
closest to its base weights while remaining more stochastic than either conventional
arm. Raising its
learning rate three-fold moves its measured KL to $0.047$, into the range the
larger arms already occupied at $10^{-6}$, and produces \tunedAllEtwo{}.

The 3$\times$~LR arms are read differently, at step 56 of their own runs, so
the two series are not interchangeable. On that step-56 measurement the 31B 3$\times$~LR
arm is at \klThirtyOneTuned{} against \klThirtyOneInc{} at step 56 of the 31B
initial run, a $16\times$ move; the 12B 3$\times$~LR arm is at \klTwelveTuned{} and
the 10$\times$~LR arm at \klPushed{}. Claims that one arm's policy move is the
largest in the study are scoped to arms measured this way, at step 56. Validation and evaluation agree at 31B, where step 56 is
simultaneously the step-matched read point and the run's validation peak; they
disagree at 12B, whose validation curve peaks at step 32 and decays through the
read point, and at 4.5B, whose peak is its final step 64. Re-evaluating that
arm there moves the score by \efourStepCheck{}. These are training-time
diagnostics from the run logs; they are not part of the evaluation and no reported
contrast depends on them.

\section{Guard failures: the \texorpdfstring{10$\times$}{10×}~LR arm and the 31B \texorpdfstring{3$\times$}{3×}~LR arm}
\label{app:degenerate}

Both higher-rate arms at 2.3B beat base on $\Us$, but only the 3$\times$~LR arm
passes the protocol's behavioural guards.

\paragraph{Why the sign differs from the RLHF scaling literature.}
\citet{hou2024rlhfscale} report the opposite sign to ours with a learned reward
model, a finite proxy whose objective gap a stronger policy optimizes sooner
\citep{gao2022overoptimization, rafailov2024daaoveropt}. Eq.~\ref{eq:reward} has no
model in the loop, so the elicitation account applies instead. RL redistributes
mass over strategies the base model already covers \citep{yue2025rlvrlimit,
akgul2026sparseselection}, bounded by coverage \citep{chen2025coverageprinciple},
which grows with size. The 10$\times$~LR arm shows the limit of that argument. The
frozen opponent pool is a second proxy in the loop, and a concession-refusing policy
that the GPT-family training buyers reward and the held-out Anthropic buyer does not
is overoptimization against that pool.

\negscaleTableGuards

The 3$\times$~LR arm passes every guard. It closes the most deals and has the
lowest seller parse-failure rate of any trained 2.3B arm, talks less
than either shared-LR trained arm, and has the narrowest buyer split in the
group. None of the guards, and none of the hand-read transcripts, distinguish it
from the shared-LR arms except
the quality of its offers.

The 10$\times$~LR arm fails three guards. It duplicates its
action array into the prose channel, once truncated, so \jsonLeak{} of its
seller turns ship a dangling \texttt{[]} or \texttt{[} to the buyer alongside the
prose, a parse-failure rate of \pushedBadParse{} over its own turns against
\tunedBadParse{} for the 3$\times$~LR arm. It also
forwards 131 unfilled prompt placeholders across its \pairedN{} episodes
(``\emph{specializing in [mention your
specialty]}''). Its prose has collapsed to a template. Of its seller turns, 53.4\%
open with the literal string ``I appreciate'' (against 17.5\% for the 3$\times$~LR arm
and 16.1\% for base), and its no-deal episodes repeat one concession-refusing
frame for all twelve of the seller's turns, issuing \textsc{counter} on 95\% of
them. Welfare is flat against the other 2.3B arms
(1.173), so the extra surplus on the deals it closes comes out of the buyer's
share. Its gain over base is \pushedGainTerra{} against \mbox{GPT-5.6-Terra} and
\pushedGainSonnet{} against Claude Sonnet~5, so its buyer split exceeds base's
by \pushedSplitVsBase{} (paired by scenario, exploratory). On $\Us$ alone it is indistinguishable from the 3$\times$~LR arm
(\pushedVsTuned{}), but not on the format and style measures above.

We report it because a reader who saw only the pooled $\Us$ column would read
\pushedGain{} as the better of the two higher-rate arms.

\paragraph{The 31B 3$\times$~LR arm fails two guards of a different kind.}
It passes every guard the 10$\times$~LR arm fails. It has a 1.000
deal rate with zero walk-aways and zero episodes at the 24-turn cap, zero bad
parses across all 8{,}400 turns, zero JSON leakage, zero unfilled placeholders,
joint welfare \welfareTunedThirtyOne{} against the 104-step arm's
\welfareIncThirtyOne{}, and a buyer-family split of \splitTunedThirtyOne{}. That
split is the narrowest in the study and is signed \emph{towards} the out-of-family
buyer, which is the opposite of an opponent-overfitting signature. Two style guards
move the wrong way. Its most common
three-word opener covers \openerTunedThirtyOne{} of seller turns against the
104-step arm's \openerIncThirtyOne{} (\openerRatioThirtyOne{}), its top three
openers together cover 63.7\%, and its repeated-opener rate is 46.7\% against
23.1\%. At 12B both
recipes sit \emph{higher} than this arm, 37.6\% for the 104-step arm
and 39.6\% for the 3$\times$~LR arm, so 30.3\% is a departure from the 31B family rather
than from the study. On the held-out car domain it opens in the \emph{buyer's}
voice on \carVoiceTuned{} of episodes against \carVoiceInc{} for the 104-step
31B arm. That second detector is the weaker of the two guards, because it does not
isolate this arm. Run over every arm (Table~\ref{tab:carvoice}) it reads 14\% to
100\%, and several arms the paper treats as clean sit near or above the 31B 3$\times$~LR
figure, including the 104-step 12B arm (76.6\%) and every E-series arm except the 10$\times$~LR arm (all above 66\%). The
31B 3$\times$~LR arm is nevertheless by far the worst of the four 31B arms, a
$5.5\times$ move against its own 104-step comparator.
Neither reading co-occurs with any 10$\times$~LR signature, neither is a
scoring artefact (the utility is earned on closed deals with rising joint
welfare), and training entropy recovers over the run rather than collapsing, so
the concentration is in the converged policy. Dialogues run
\turnsTunedThirtyOne{} turns against \turnsIncThirtyOne{}, \turnsCutThirtyOne{}
shorter, and 54.1\% of its closing turns name
the concession being traded, against 18.0\% for the 12B 3$\times$~LR arm
(a cross-size comparison; this detector was not run on the 104-step 31B arm).

These are real costs of the 3$\times$~LR at 31B against an arm it does not measurably
beat (interval $-0.014$ to $+0.020$). The 31B 3$\times$~LR arm also has the largest
step-56 KL to base in the study (\klThirtyOneTuned{}) and passes all four
prespecified guards, while the 10$\times$~LR arm has a smaller KL (\klPushed{})
and games the opponent pool. In our runs the size of the policy change did not
track Goodharting, so we judge each arm on its behavioural guards.

\paragraph{Four guards were prespecified; the rest of the list, and its partition,
came later.} The protocol prespecifies four anti-Goodhart guards: deal rate must
not collapse while $\Ut$ rises, the bad-parse fraction must stay flat, mean turns
must not saturate at the cap, and five transcripts per RL'd arm per domain family
are read by hand. Welfare appears there as a secondary metric, not as a guard. The
other checks reported here (walk-aways, JSON leakage, template-tag leaks, unfilled
placeholders, the buyer-family split, opener concentration and held-out role
confusion) were added as successive evaluation rounds surfaced them and entered through the
deviation log rather than the frozen protocol. Opener concentration and held-out
role confusion in particular were recorded as guard \emph{failures} for the 31B 3$\times$~LR arm only after its readings were in hand; the 12B
3$\times$~LR arm's comparable
readings, evaluated earlier, had been logged as a property of the recipe. The split of the list into
``utility and format'' guards, which the 31B 3$\times$~LR arm passes, and ``style''
guards, which it fails, was drawn the same way, after seeing this arm. Of the
guards fixed in advance, this arm passes all four. The two it fails, opener
concentration and held-out role confusion, are post-hoc style guards, and neither
measures what the seller extracts or whether the buyer can parse its turns.

\begin{table}[h]
  \centering
  \caption{Buyer's-voice opener rate on the held-out car domain, one detector over
  every evaluated arm ($n = \heldoutN{}$ car episodes each). The detector separates
  arms weakly and separates \emph{sizes} strongly, so it supports the within-31B
  comparison in the text and not a cross-size one. Absolute rates depend on the
  detector's breadth, so only within-table comparisons are meaningful.}
  \label{tab:carvoice}
  \small
  \setlength{\tabcolsep}{4pt}
  \begin{tabular}{lccccc}
    \toprule
    Arm & base & shared LR initial & shared LR, 104 steps & 3$\times$~LR & 10$\times$~LR \\
    \midrule
    E2B (2.3B) & 87.5\% & 80.5\% & 66.4\% & 100.0\% & 37.5\% \\
    E4B (4.5B) & 89.1\% & 89.8\% & 86.7\% & 92.2\% & --- \\
    12B        & 46.9\% & 65.6\% & 76.6\% & 61.7\% & --- \\
    31B        & 27.3\% & 18.0\% & 14.1\% & \textbf{78.1\%} & --- \\
    \midrule
    \multicolumn{6}{l}{\emph{Frontier reference sellers}: Claude Sonnet~5 57.8\%, GPT-5.6-Terra 18.8\%} \\
    \bottomrule
  \end{tabular}
\end{table}

\section{The quoted episode}
\label{app:example}

The exchange quoted in Section~\ref{sec:method} is turns 6 and 7 of one
evaluation episode on the held-out car domain, the 12B 3$\times$~LR seller against
GPT-5.6-Terra, seed 5000053.

Before the seller's turn 7 the ledger held three terms (model Sedan, interior Premium,
price 40). Terms named in a \textsc{counter} overwrite the ledger, which holds
every term named so far, so after the turn it holds five of the domain's ten,
with price unchanged and service and accessories added. The five terms never
named each contribute $0$ to both parties' $U$, though they can still add to
normalized utility (Section~\ref{sec:method}). On turn 10 the buyer kept the
annual service but set accessories to false, and the seller accepted that ledger
on turn 11. Evaluation ignores terms attached to an \textsc{accept}, so the terms
the seller listed with its own were dropped and the deal was scored on that
ledger.

\section{Serving cost of each arm}
\label{app:efficiency}

Table~\ref{tab:efficiency} sets each arm's score beside the hardware it needs to
serve. The 31B arms need four 48\,GB accelerators to hold a 16k-context
KV cache while every smaller arm fits on one, and the 12B 3$\times$~LR seller is
the smallest arm whose paired contrasts against both frontier reference sellers
exclude zero (Section~\ref{sec:results}). We did not meter
GPU-seconds, energy, or throughput, so the table compares serving capacity only.

\negscaleTableEfficiency

\section{Training recipe}
\label{app:recipe}

Every arm runs the recipe below, identical up to the learning-rate level of the
factorial ($10^{-6}$ shown; the 3$\times$ and 10$\times$~LR arms scale the
peak and the cosine floor by the same factor, to floors of $3{\times}10^{-7}$
and $10^{-6}$), in the shape now standard for open
post-training pipelines \citep{lambert2024tulu3}. The only per-size differences
\emph{in the training recipe} are
the hardware-forced ones in Table~\ref{tab:topology}, none of which enters the
objective, the data, or any optimizer hyperparameter; the sizes' own chat
templates also differ, which we keep by design (Appendix~\ref{app:template}).

\begin{itemize}\itemsep2pt
  \item \textbf{Algorithm.} GRPO \citep{shao2024deepseekmath} with reward
  normalization and a leave-one-out baseline; no learned value model. PPO clip
  $0.2/0.2$, token-level loss. KL penalty $0.04$ ($k_3$ estimator) against the
  initial policy, applied to the loss and \emph{not} folded into the reward.
  \item \textbf{Batching.} $8$ rollouts per scenario (the GRPO group) over $16$
  scenarios per step $= 128$ episodes/step. The initial run is $64$ steps
  ($8{,}192$ episodes); the continued steps add $40$ or $48$ more to reach
  $104$ combined steps ($13{,}312$ episodes). Validation at step $0$ and every
  $8$ steps.
  \item \textbf{Optimizer.} Fused Adam, learning rate $10^{-6}$,
  $\beta = (0.9, 0.95)$, weight decay $0.01$, bf16 moments and master weights;
  linear warmup over $5$ steps into a cosine decay to $10^{-7}$. The continued
  steps restart this schedule from their own warmup rather than resuming the parent
  cosine, so a 104-step arm runs two schedules back to back.
  \item \textbf{Sequences.} $16{,}384$-token context, $2{,}048$-token prompt cap,
  $768$ new tokens per turn, up to $13$ rollout turns in the driver, one above
  the $12$ seller turns the round cap allows, so the environment rather than the
  driver ends every episode.
  \item \textbf{Sampling.} Temperature $1.0$, top-$p$ $0.95$, top-$k$ $64$,
  thinking mode disabled. Evaluation uses the same values for every self-served
  seller, and the harness warns at startup if they drift.
  \item \textbf{Reward plumbing.} NeMo-RL's driver sums per-step rewards into the
  episode return and exposes no post-batch hook, so we front-load the floor: an
  episode emits $-\lambda$ on its first step and Eq.~\ref{eq:reward}
  $+\,\lambda$ on its terminating step, making the sums exact and leaving a
  truncated episode with the floor.
  \item \textbf{Environment.} $\lambda = 0.3$, $g_{\mathrm{fmt}} =
  g_{\mathrm{len}} = 0.05$, $12$-round cap, one training seed per size.
\end{itemize}

\begin{table}[h]
  \centering
  \caption{Per-size parameter counts, serving topology and relative training
  compute at matched optimizer steps. Tensor-parallel degree is capped by the
  checkpoint's key/value head count and by device memory, not chosen.}
  \label{tab:topology}
  \small
  \begin{tabular}{lrrrr}
    \toprule
    & E2B & E4B & 12B & 31B \\
    \midrule
    Effective parameters & 2.3B & 4.5B & 11.95B & 30.7B \\
    Total parameters & 5.1B & 8.0B & 11.95B & 30.7B \\
    Train tensor-parallel degree & 1 & 1 & 1 & 4 \\
    Train micro-batch size & 1 & 1 & 1 & 1 \\
    Generation memory fraction & 0.45 & 0.40 & 0.35 & 0.30 \\
    Eval serving accelerators (48\,GB) & 1 & 1 & 1 & 4 \\
    Relative FLOPs/step (effective params) & $1.0\times$ & $2.0\times$ & $5.2\times$ & $13.3\times$ \\
    Relative FLOPs/step (total params) & $1.0\times$ & $1.6\times$ & $2.3\times$ & $6.0\times$ \\
    \bottomrule
  \end{tabular}
\end{table}

\section{FLOP accounting for the matched-step budget}
\label{app:flops}

Training compute for one GRPO step is approximately
$C \approx 2 N D_{\mathrm{gen}} + 6 N D_{\mathrm{train}}$, where $N$ is the
parameter count entering the forward pass and $D$ are generated and trained token
counts. Every arm runs the same $128$ episodes per step over the same scenario
distribution, with the same context cap, the same $768$-token per-turn budget and
the same group size, so $D_{\mathrm{gen}}$ and $D_{\mathrm{train}}$ differ across
arms only through realized dialogue length. Measured mean episode length at
evaluation ranges from $9.9$ to $12.8$ turns across the base and shared-recipe
Gemma arms that this accounting covers, so their token counts differ by at most
about $30\%$ while $N$ differs by $13\times$. Per-step compute is
therefore approximately proportional to $N$, giving the ratios in
Table~\ref{tab:topology}. Across all \numArms{} evaluated arms the range is much
wider, \turnsTunedThirtyOne{} turns for the 31B 3$\times$~LR arm to $14.4$ for the
10$\times$~LR arm, a factor of about two. Those arms vary the learning rate rather
than the step budget, so they are not part of the matched-step comparison, and the
$30\%$ figure does not cover them.

We report both parameter conventions because they bracket the truth for the
E-series. Forward and backward passes are dominated by the effective parameter
count, while the optimizer state and the update touch all parameters, so the
E-series arms sit between the two rows. Under either convention the 31B arm
received the most training compute at 104 steps, which is the reason
Section~\ref{sec:discussion} splits the rate question from the compute question
rather than
claiming a compute-matched design. The 31B arm's rise from $0.564$ at $56$ steps
to \armThirtyOneRlPt{} at $104$ could in principle be a restart artefact rather
than extra training, and we cannot separate the two. Every arm restarts its
schedule identically, and the E-series continuations move little, but at those
sizes the shared rate barely trains at all, so their null is weak evidence
against a restart effect. We did not meter wall-clock
GPU-hours per arm, because the runs were interleaved on shared, partly preemptible
capacity, so we do not report measured FLOPs.

\section{Chat-template divergence across sizes}
\label{app:template}

The four sizes ship different chat templates. The divergence is present in the
current published repositories of all four sizes, not an unsynced artefact, and
both variants render balanced turn tags. With thinking mode
disabled (the setting every arm runs) the conventional checkpoints' template appends an empty
thought-channel prologue to the generation prompt that the E-series template does
not, so E-series generation prompts are $10$ tokens against the conventional sizes' $14$:
\begin{center}
\begin{tabular}{ll}
  E2B, E4B & \texttt{\ldots<|turn>model\textbackslash n} \\
  12B, 31B & \texttt{\ldots<|turn>model\textbackslash n<|channel>thought\textbackslash n<channel|>} \\
\end{tabular}
\end{center}
We use each model's own published template. Grafting the conventional template
onto the E-series would condition those models on a prefix pattern absent from
their own instruction tuning, which is a larger and less predictable intervention
than the difference it removes. Each
arm's base counterpart is served with the same template as its trained
counterpart, so the base-versus-RL contrast, the quantity the paper reports,
is unaffected. Because the rollout driver tokenizes environment observations
raw, the environment must emit the buyer's turn tags itself; we derive them per
arm from that arm's tokenizer by differencing renders of a test dialogue, and
assert the extraction round-trips on a multi-turn history before training starts.
Every served prompt was verified by tokenizing and detokenizing it before the arm
was evaluated, and no arm produced a template-tag leak into buyer-visible text.

\section{Evaluation protocol deviations}
\label{app:deviations}

Several protocol details are held out of Section~\ref{sec:setup} for space.
The primary family is Holm-corrected on bootstrap p-values from the same
resampling. The 4.5B, 12B and 31B gains are significant with or without
correction ($p \le 1.4{\times}10^{-4}$ against Holm thresholds of
$0.0125$--$0.025$), and the 2.3B gain is not ($p = 0.91$).
Evaluation seeds come from a reserved block three orders of magnitude above every
training seed. At $10{,}000$ resamples the interval endpoints carry up to about
$\pm0.001$ of Monte Carlo error. The two contrasts whose
lower bound sits inside that band (\diffThirtyOneTwelvePt{} within the conventional
pair and \tunedStepThirtyOneTwelvePt{} between the 12B and 31B 3$\times$~LR arms) are therefore re-run at
$10^{5}$ resamples, where the achieved endpoint error is \mcError{} and the two
lower bounds settle at $+0.0008$ and $+0.0028$. Both survive Monte Carlo noise and
neither survives it by more than three thousandths of $\Us$, so both are reported
as positive and only marginally distinguishable from zero, unadjusted. No pair
dropped an episode in any contrast reported in this paper. Mean turns counts individual turns, two per round, so the 12-round cap is
24. Two guards hold on every arm in the main table: the seller's prose reaches
the buyer, and no chat-template tag leaks into buyer-visible text.

The evaluation protocol was frozen before any arm ran, and every departure from it
is logged against the arms it applies to. The material ones:

\begin{itemize}\itemsep2pt
  \item \textbf{A harness revision partway through.} An earlier version of the
  harness silently stripped the seller's prose before the buyer saw it, gave the
  seller half its trained horizon, and decoded at a different temperature with
  the top-$p$/top-$k$ truncation not applied. The correction moved arm-level
  $\Us$ by up to about $0.4$, far larger than any contrast in this paper, which
  is why we treat evaluation fidelity as first-order. Every
  number in this paper comes from the corrected harness, and no result from the
  earlier one is reported or mixed into any table or figure.
  \item \textbf{The 31B 104-step arm is not its run's best checkpoint.} The
  matched-step rule selects step 48 of the 31B continued steps (validation $0.645$),
  below both the run's validation peak at step 56 ($0.676$) and its final step
  ($0.659$). The arm still produced the largest gain under the shared recipe, so
  the reported \gainAllThirtyOnePt{} is the conservative choice rather than a
  favourably selected one.
  \item \textbf{Checkpoint substitutions.} Two initial runs lost their final
  checkpoint to an interrupted weight upload, so the 4.5B and 31B initial arms are
  read at step 56 rather than 64 (validation $0.195$ versus $0.198$, and $0.600$
  versus $0.633$). On validation both substituted checkpoints score lower, so
  the substitution is unlikely to flatter the reported gain.
  \item \textbf{The 2.3B 3$\times$ and 10$\times$~LR arms are read at step 56}, owing to
  the same upload failure, while the 2.3B shared-LR arm they are compared with is
  read at step 64, so both are 8 steps short. On the 3$\times$~LR arm the
  effect is neutral, its validation curve being flat ($0.4084$ at 56 against
  $0.4109$ at 64). On the 10$\times$~LR arm step 64 scores higher ($0.659$ against
  $0.621$), so \pushedGainPt{} may understate it. The case against that arm rests on its
  behaviour (Appendix~\ref{app:degenerate}).
  \item \textbf{The 12B 3$\times$~LR arm is read at step 56} by choice, to match
  the other 3$\times$~LR arms, while the 12B shared-LR arm it is compared with is
  read at step 64. Its validation curve peaks at step 32 ($0.481$) and falls to
  $0.449$ at step 56 and $0.368$ at step 64, so a step-64 read would probably
  score lower.
  \item \textbf{None of the five 3$\times$ and 10$\times$~LR arms is in the frozen arm
  matrix}, which names one learning rate per size. Seeds, buyers, domains, sampling,
  horizon and analysis follow the protocol exactly, so all five pair
  scenario-for-scenario with every other arm.
  \item \textbf{The 10$\times$~LR Sonnet~5 cell did not clear the pre-run pilot
  gate}, returning 3 bad parses in 54 turns where the gate requires zero. We ran
  it regardless, because that arm's format stability is what it was
  commissioned to measure and aborting would have discarded the finding.
  \item \textbf{Two schedules per 104-step arm}, as in
  Appendix~\ref{app:recipe}, with 40 or 48 continued steps
  depending on where the parent run stopped.
  \item \textbf{Serving topology.} The 31B arms are served with four
  accelerators rather than two; at two the KV cache cannot hold the 16k context.
  The serving argument string is otherwise identical across arms, and the 31B
  base arm it is contrasted against is served the same way.
  \item \textbf{Nash distance and Pareto efficiency are not reported.} We did not
  compute them for this paper; joint welfare is the only efficiency proxy.
  \item \textbf{Strategy descriptors were added at camera-ready}, in response to
  review: Table~\ref{tab:strategy}, its intervals in Table~\ref{tab:strategyfull},
  and the closing-turn shares in Section~\ref{sec:results}. All are computed from
  the frozen evaluation transcripts. The $\Delta$ columns use the same pairing and
  seed-cluster bootstrap as every other contrast, restricted to the scenarios both
  arms closed. None was prespecified, and all are exploratory.
  \item \textbf{The buyer-split interval was added at camera-ready.} The
  10$\times$~LR arm's gain against each buyer and their difference
  (Appendix~\ref{app:degenerate}) use the same pairing and bootstrap.
  \item \textbf{Transient infrastructure failures} produced error rows in several
  runs. Every affected episode was re-run, and the results underlying this paper
  contain zero error rows.
  \item \textbf{The prespecified held-out-panel extension was not run}, as
  Section~\ref{sec:setup} states.
\end{itemize}

\negscaleTableStrategyFull

\end{document}